\documentclass{article} 
\usepackage{the-nvidia-pilotnet-experiments}
\usepackage{xcolor}
\usepackage[]{hyperref}
\hypersetup{
  colorlinks=true,
  allcolors=blue,
  linkcolor=black
}
\usepackage[labelfont=bf]{caption}
\usepackage[labelfont=bf]{subcaption}
\usepackage{amsmath}
\usepackage{url}
\usepackage{graphicx}           
\usepackage{tabu}               
\usepackage[nopostdot,style=super,nonumberlist]{glossaries}
\makeglossaries
\usepackage{nips20submit,times}
\nipsfinalcopy
\title{The NVIDIA PilotNet Experiments}
\author{
  \hspace{-1.1em}       
  \begin{minipage}{0.85\textwidth}\rm\center
    \begin{centering}
    Mariusz~Bojarski, Chenyi~Chen, Joyjit~Daw, Alperen~De\u{g}irmenci,
    Joya~Deri, Bernhard~Firner, Beat~Flepp, Sachin~Gogri, Jesse~Hong,
    Lawrence~Jackel, Zhenhua~Jia, BJ~Lee, Bo~Liu, Fei~Liu, Urs~Muller,
    Samuel~Payne, Nischal~Kota~Nagendra~Prasad, Artem~Provodin,
    John~Roach, Timur~Rvachov, Neha~Tadimeti, Jesper~van~Engelen,
    Haiguang~Wen, Eric~Yang, and Zongyi~Yang
    \end{centering}
  \end{minipage}
  \\
  \\
  NVIDIA Corporation \\
  101 Crawfords Corner Road \\
  Holmdel, NJ 07733 \\
}
\newacronym{acc}{ACC}{adaptive cruise control}
\newacronym{alvinn}{ALVINN}{Autonomous Land Vehicle In a Neural Network}

\newacronym{can}{CAN}{Controller Area Network}
\newacronym{ces}{CES}{Consumer Electronics Show}
\newacronym{ci}{CI}{continuous integration}
\newacronym{cmu}{CMU}{Carnegie Mellon University}

\newacronym{darpa}{DARPA}{Defense Advanced Research Projects Agency}
\newacronym{dave}{DAVE}{DARPA Autonomous Vehicle}
\newacronym{dnn}{DNN}{deep neural network}

\newacronym{gps}{GPS}{Global Positioning System}
\newacronym{gpu}{GPU}{Graphical Processing Unit}
\newacronym{gtc}{GTC}{\gls{gpu} Technology Conference}

\newacronym{imu}{IMU}{inertial measurement unit}

\newacronym{lagr}{LAGR}{Learning Applied to Ground Robots}

\newacronym{mapa}{MAPA}{model affinity to perturbation artifacts}
\newacronym{mdbf}{MDBF}{mean distance between failures}

\newacronym{obd}{OBD}{on-board diagnostics}

\newacronym{qa}{QA}{quality assurance}

\newacronym{rgb}{RGB}{red green blue}
\newacronym{rms}{RMS}{root mean square}
\newacronym{roi}{ROI}{Region of Interest}

\begin{document}
\maketitle
%
%
%
%
%
%
%
%

\begin{abstract}
Four years ago, an experimental system known as PilotNet became the
first NVIDIA system to steer an autonomous car along a roadway. This
system represents a departure from the classical approach for
self-driving in which the process is manually decomposed into a series
of modules, each performing a different task. In PilotNet, on the
other hand, a single \gls{dnn} takes pixels as input and produces a
desired vehicle trajectory as output; there are no distinct internal
modules connected by human-designed interfaces. We believe that
handcrafted interfaces ultimately limit performance by restricting
information flow through the system and that a learned approach, in
combination with other artificial intelligence systems that add
redundancy, will lead to better overall performing systems. We
continue to conduct research toward that goal.\vspace{1mm}

This document describes the PilotNet lane-keeping effort, carried out
over the past five years by our NVIDIA PilotNet group in Holmdel, New
Jersey. Here we present a snapshot of system status in mid-2020 and
highlight some of the work done by the PilotNet group.\vspace{1mm}

PilotNet's core is a multi-layer neural network that generates lane
boundaries and the desired trajectory for a self-driving vehicle.
PilotNet works downstream of systems that gather and preprocess live
video of the road. A separate control system, when fed trajectories
from PilotNet, can steer a vehicle. All systems run on the NVIDIA
DRIVE\texttrademark~AGX platform.\vspace{1mm}

Substantial infrastructure was built by numerous NVIDIA teams to
support PilotNet. A huge corpus of training data has been collected,
filtered, and annotated. A high-fidelity simulation based on real
video recordings tests the performance of the resulting neural
networks. A software pipeline automates much of the neural-net
training process which runs on a large cluster of remote computing
resources. In addition, an in-car status monitor allows human safety
operators to view PilotNet trajectory output while an autonomous
vehicle is in operation.\vspace{1mm}

While the basic principles that govern our network's supervised
learning are straightforward, numerous enhancements have been created
to achieve good performance.\vspace{1mm}

Using a single front-facing camera, the best current version of
PilotNet can steer an average of about 500 km on highways before a
human takeover is required. We note that this result was obtained
without using lidar, radar, or maps and is thus not directly
comparable to most other published measurements. We suggest that an
approach like PilotNet can enhance overall safety when employed as a
component in systems that use additional sensors and maps.

\subsection*{A Guide to the Reader}
Sections~\ref{sec-intro}, \ref{sec-historical}, and \ref{sec-overview}
provide a historical perspective and overview of our technical approach.
Sections~\ref{sec-collection} to~\ref{sec-multires} describe data
collection, data preparation, network training, and performance metrics
in a fair amount of technical detail. Section~\ref{sec-tuning} discusses
a case where optimizing results in simulation led to degraded on-the-road
driving and describes how we discovered the root cause and corrected the
problem. Section~\ref{sec-lessons} presents some lessons learned and
Section~\ref{sec-conclusions} is a short conclusion.\vspace{1mm}

Before the list of references is a list of links to representative
videos. We suggest that the reader who seeks an overview, focus on
Sections~\ref{sec-intro}--\ref{sec-overview}, \ref{sec-perf-history},
and \ref{sec-lessons}, as well as the short videos.
\end{abstract}

\printglossary[title={Abbreviations}]

\section{Introduction}
\label{sec-intro}
PilotNet is a research software system that guides an autonomous car
along a roadway. In PilotNet, a single Deep Neural Network (DNN) takes
pixels as input and produces a desired vehicle trajectory as
output. This system represents a departure from the classical approach
for self-driving in which the process is manually decomposed into a
series of modules, each performing a different task.

A guiding philosophy in building PilotNet has been the desire to
minimize the use of both handcrafted features and predetermined
modularization. We strove to create a learned system that avoids
imposing human-determined boundaries and interfaces between successive
elements of PilotNet's processing sequence. We avoided such
modularization because we have observed that manually engineered
interfaces can fail to capture all required information and fail to
transfer crucial information from one stage of the process to the
next. When combined with diverse and redundant systems, we expect that
PilotNet will provide enhanced safety.

\subsection{Learned Neural Networks versus Handcrafted Rules}
Some of the first applications of artificial intelligence were expert
systems in which an algorithm works its way through a series of
if-then-else decisions to go from input data to some output. Usually
these rules are handcrafted, often with insight provided by a human
subject matter expert. An advantage of such a system is that its
behavior can be explained since the rules can be expressed in plain
language or as mathematical expressions. Yet there are many
circumstances where a handcrafting approach fails; this is
particularly true in tasks where humans can't elucidate the rules,
such as tasks that involve human perception. As an example,
image-based object recognition has been attempted for decades using
handcrafted feature extraction, but with only limited success.

An alternative to handcrafting emerged in the late 1980s with the
popularization of neural networks in which perception problems were
solved by using large collections of labeled examples (\eg, image /
label pairs) to adjust the network weights. Among the early examples
of this learned approach were the convolution networks developed by
LeCun and colleagues for hand-printed digit recognition in
1989~\cite{lecun-89e}. Essentially the same approach, only scaled to
much larger networks and to many more training examples, is used today
in most state of the art recognition systems.

For many people, a drawback of large neural networks is that it is
usually impossible to fully understand how the network produces its
output. In effect, large networks can be considered extremely complex
rules-based systems in which the rules are learned. The complexity of
these networks is so great that they exceed the ability of humans to
comprehend their internal workings. We believe this complexity
reflects an intrinsic property of the task and that we need to accept
that high performance inevitably comes at the price of fine-grained
explainability. Based on our experience, and observing developments
in the field as a whole, we are convinced that for tasks like
autonomous driving, a handcrafted rules approach is unlikely to
achieve the performance of a learned system that has access to big
data.

\begin{figure}[htb]
  \hfil\includegraphics[width=0.7\textwidth]{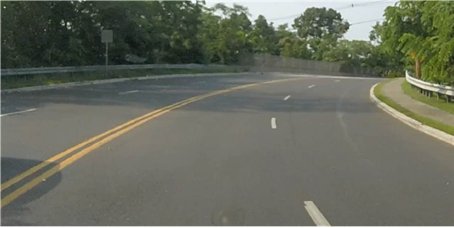}

  \hfil\includegraphics[width=0.7\textwidth]{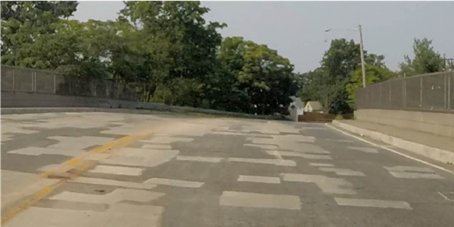}
  \caption{A highway bridge near our Holmdel office. A human designed
    rules-based system could navigate on the top road section, but
    would struggle on 50\m\ along the same road (bottom).}
  \label{fig-hwy-bridge}
\end{figure}

\begin{figure}[htb]
  \hfil\includegraphics[width=0.8\textwidth]{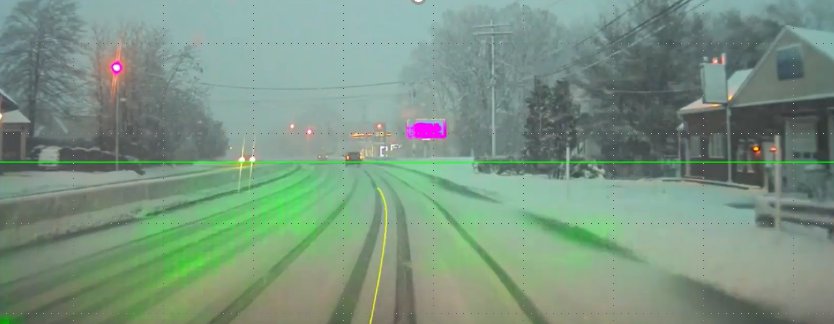}
  \caption{A snowy road in New Jersey in 2018. PilotNet was able to
    drive here; it would have been an extreme challenge for a
    handcrafted rules-based system. The green highlighted regions were
    most salient in determining the proper trajectory (yellow curve).}
  \label{fig-snowy-road}
\end{figure}

Large neural networks can learn behaviors that are extremely
challenging or impossible for human-engineered rules-based
systems. Consider a navigation system operating in situations like
those in Figures~\ref{fig-hwy-bridge} (bottom)
and~\ref{fig-snowy-road} where the visual cues are not clear. While a
system with human-crafted rules could do a passable job navigating
over the bridge in Figure~\ref{fig-hwy-bridge} (top), such a system
would probably fail 50\m\ further down the same road where the paving
has been repeatedly patched as shown in Figure~\ref{fig-hwy-bridge}
(bottom).

As another example, consider how would one create the proper rules for
driving on the snowy road in Figure~\ref{fig-snowy-road}. PilotNet,
which will be described in this document, does a credible job in these
conditions.

\section{A Historical Perspective}
\label{sec-historical}
\begin{figure}[htb]
  \hfil
  \includegraphics[width=0.494\textwidth]{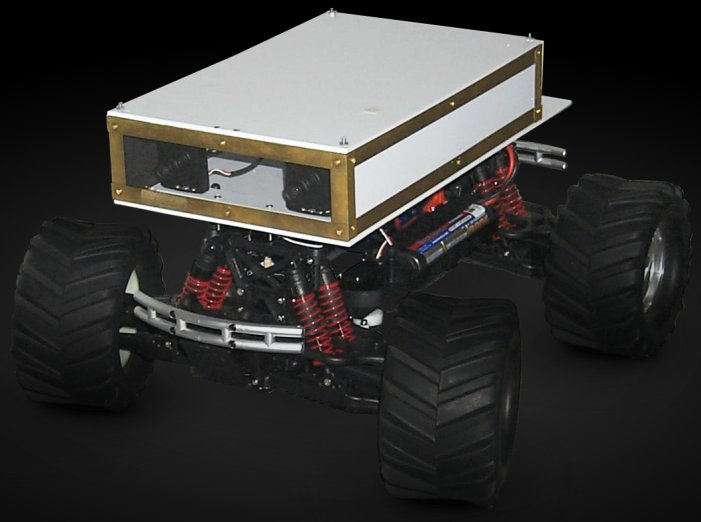}
  \includegraphics[width=0.48\textwidth]{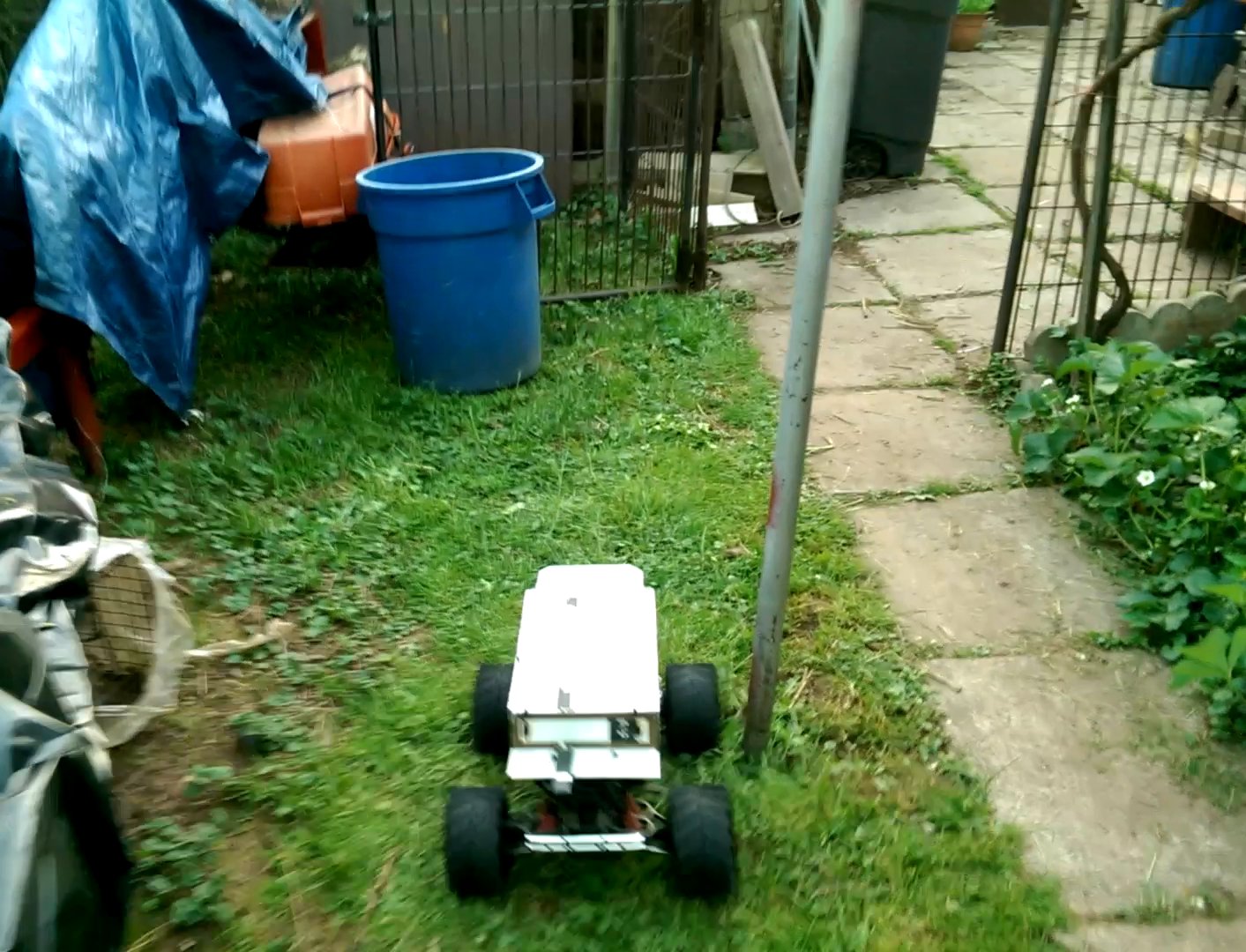}
  \caption{The robot used in the \gls{dave} project is a modified
    50-cm long model car platform controlled by a remote computer. The
    vehicle's main sensors are two front cameras with 320\x240 pixel
    resolution. No explicit stereo algorithm was applied. The two
    camera images were simply fed to the neural network as separate
    planes.}
  \label{fig-dave}
\end{figure}

\glsunset{dave}
PilotNet's roots go back to 2003. At that time the \gls{darpa} funded a
six-month exploratory project to see if it was possible to learn a
complete control loop from sensors to actuators. The target
application was autonomous off-road navigation for mobile ground
robots; the project was named ``\gls{dave}'' for \gls{darpa} Autonomous Vehicle
Experiment~\cite{LMB*05}. Figure~\ref{fig-dave} shows the DAVE robot.

\gls{dave} learned from steering information captured while a human
remote-controlled the vehicle. This end-to-end learning approach is
similar to what Dean Pomerleau created at \gls{cmu} in 1989 in his
\gls{alvinn} project~\cite{Pom89}. In the intervening 14~years there
were tremendous advances in computing hardware, in camera technology,
and in the understanding of how to train large neural networks, so it
was possible to advance beyond \gls{alvinn}. In particular, during the
1990s, the convolutional neural networks, pioneered by LeCun and his
colleagues at Bell Labs~\cite{lecun-89e} provided a huge performance
jump in image analysis.

\begin{figure}[htb]
  \hfil
  \includegraphics[width=\textwidth]{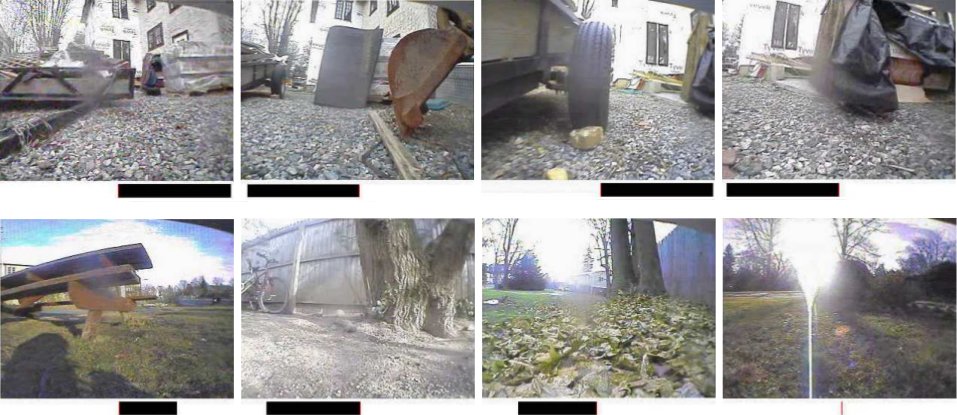}
  \caption{Snapshots from \gls{dave}'s left camera taken while the
    robot drove through various environments. The black bar beneath
    each image indicates the steering angle produced by the
    system. Top row: four successive snapshots showing the robot
    navigating through a narrow passageway between a trailer, a
    backhoe, and some construction material. Bottom row, left: narrow
    obstacles such as table legs and poles (left), and solid obstacles
    such as fences (center-left) were easily detected and
    avoided. Highly textured objects on the ground did not detract the
    system from the correct response (center-right). One scenario
    where the vehicle frequently made wrong decisions is when the sun
    is in the field of view: the system seemed to systematically drive
    towards the sun whenever the sun is low on the horizon (right).}
  \label{fig-dave-snapshots}
\end{figure}

In the \gls{dave} project, \gls{darpa} Program Manager Larry Jackel's
former Bell Labs and AT\&T Research colleagues, Yann LeCun, Urs Muller,
Beat Flepp, and Eric Cosatto demonstrated that a single convolutional
neural network can be trained to drive a robot vehicle. The \gls{dave}
vehicle was a small radio-controlled robot car. \gls{dave} was able to
navigate through a home construction site littered with debris for
tens of meters. While \gls{dave} could not drive long distances
without crashing, nonetheless, it held the promise of a breakthrough
in autonomous driving. Images from one of \gls{dave}'s cameras are
shown in Figure~\ref{fig-dave-snapshots}.

In 2004 \gls{darpa} was focusing on off-road driving. To benchmark the
state of the art, \gls{darpa} sponsored a Grand Challenge. Initially
the plan was for competing autonomous vehicles to drive across the
Mojave Desert from near Los Angeles to near Las Vegas. For logistics
reasons the route was simplified to mostly follow a power line
right-of-way from Barstow, California to Primm, Nevada. In this first
Grand Challenge even the best participants only traveled a few miles
before getting stuck. As far as we know, machine learning played
little or no role in this event.

In contrast, and also in 2004, based on the success of the 6-month
\gls{dave} project, \gls{darpa} started the \gls{lagr}~\cite{JHK*07}
program. In \gls{lagr}, teams from top universities and research
organizations competed to drive over ever more challenging
courses. Teams were encouraged to use learning as much as possible in
their navigation and control software. Over the 4-year lifetime of the
program considerable progress was achieved.

In 2005 \gls{darpa} ran a second Grand Challenge. This time a team led
by Sebastian Thrun of Stanford won, completing the course. Part of the
Stanford software stack used learning.

In 2007 \gls{darpa} ran the Urban Challenge in which autonomous
vehicles competed on a course in an abandoned army base that had local
streets and many buildings. In addition to avoiding other competitors,
the vehicles had to contend with traffic generated by a fleet of Ford
Taurus cars conservatively driven by professional drivers. Navigation
in the Urban Challenge was simplified with \gls{darpa} providing dense
way-points along the route and teams having access to differential
\glspl{gps}. This was also the first event in which Velodyne LIDAR was
widely deployed, giving users 3D information that proved crucial for
avoiding both moving and stationary obstacles. The Urban Challenge was
won by a team from CMU led by Chris Urmson.

The success of the Urban Challenge led to today's commercial efforts
to build self-driving cars. Many of the key researchers in the
\gls{darpa} Challenges are leading industry efforts today and many of
these efforts use ideas generated by the Challenges.

The years following the \gls{darpa} Challenges witnessed an explosion
of interest in Deep Learning. Convolutional neural networks went from
a shunned technology to mainstream. The realization that \glspl{gpu}
could vastly accelerate both the learning process and the execution
(now called ``inference'') of a trained neural network has played a
major role in this explosion and has encouraged the idea that
practical self-driving cars are possible.

In 2015 Urs Muller and Beat Flepp joined NVIDIA to help create
software systems for self-driving cars. Building on the work they did
for \gls{darpa} in the \gls{dave} project they began to create a
system now known as PilotNet.

\section{PilotNet Overview}
\label{sec-overview}

\subsection{Evolution of PilotNet}
The original PilotNet system was created by NVIDIA in Holmdel in
2015-–2016 and is described in some depth in a paper by Bojarski et
al~\cite{dave2-2016}. PilotNet drew on ideas developed by
Pomerleau~\cite{Pom89} (\gls{alvinn}) in 1989, and later by LeCun et
al~\cite{LMB*05} (\gls{dave}) in 2006. In both \gls{alvinn} and
\gls{dave}, recorded images of the view from the front of a vehicle
were simultaneously logged along with the steering commands of a human
driver creating input-target pairs. These pairs were then used for
training a multi-layer neural network.

Both \gls{alvinn} and \gls{dave} were limited by the computing
resources available at the time of their creation. \gls{alvinn}
featured a very small (by today's standards) fully-connected neural
network (30 hidden units) with low-resolution input images coupled
with coarse range-finder input. \gls{dave} used a convolutional neural
network~\cite{lecun-89e} with inputs from a pair of cameras. While
\gls{dave}'s cameras did not explicitly calculate depth from stereo,
the use of two cameras meant that depth from stereo could be
learned. Both \gls{dave} and \gls{alvinn} were far from perfect, but
they both represented advances over contemporaneous rule-based
autonomous navigation systems.

Muller and Flepp joined NVIDIA in February of 2015 forming our core
PilotNet team. We realized that creating a \gls{dave}-like system
that runs on real cars on public roads entails great technical risk
and success was far from assured. Open questions included:

\begin{itemize}
  \item[$\Rightarrow$] Can an end-to-end approach scale to production?
  \item[$\Rightarrow$] Does the training signal obtained by pairing
    human steering commands with video of the road ahead contain
    enough information to learn to steer a car autonomously?
  \item[$\Rightarrow$] Can an end-to-end neural network outperform more traditional
    approaches?
\end{itemize}

We immediately set about to see if the DAVE concepts could be applied
to on-road driving. With the addition of new hires and a pair of
summer interns we began collecting driving data.

\begin{figure}[htb]
  \hfil
  \includegraphics[width=0.6\textwidth]{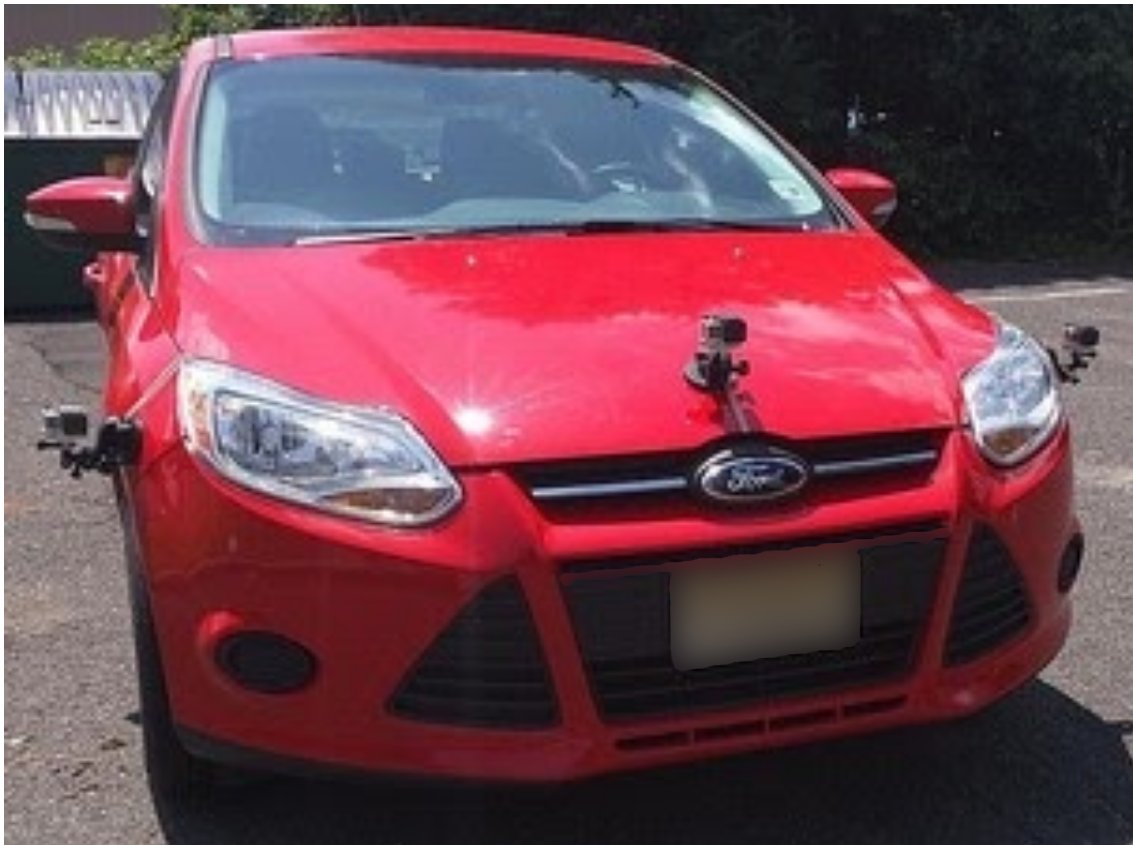}
  \caption{An early data collection car. Suction cups held GoPro
    cameras on the front of the vehicle, left, center, and right. The
    off-center cameras facilitated augmenting the training data.}
  \label{fig-focus1}
\end{figure}

In the first ``hello world'' experiment, GoPro cameras were mounted on
the hood of a car using suction cups; one camera on the left side, one
on the right, and one centered. Video was recorded from each camera as
the car was driven along a road. A network was tasked to classify
images as coming from either the left, center, or right camera. This
task was learned successfully, giving confidence that more complex
learning was possible.

In the next experiments, we utilized a stock Ford Focus, since this
model's \gls{can} codes for steering and speed were known to us. We
could then use a consumer Bluetooth \gls{obd} dongle paired with a
smartphone.

The cameras on our data collection car were moved to a roof rack
providing better, consistent camera calibration. The \gls{can} data
was time-stamped with the phone's network time. To synchronize the
phone with the cameras, the display of a clock app on the phone was
placed in front of the cameras at the start of a data collection run
so that the cameras recorded the phone time. We found that the GoPros
stayed synchronized with the phone time for the course of a full data
collection run, typically about one hour.

In creating \gls{alvinn}, Pomerleau noted the need
for augmenting the training set so that the autonomous system could
drive even if the vehicle strayed beyond the range of the training
data. Thus the training set was augmented with transformed images that
showed what the vehicle camera might record if it had drifted away
from the center of the road. These images were coupled with a steering
angle that would direct the vehicle back to the center.

In PilotNet, the left and right cameras were used to provide augmented
training data. These cameras recorded a view that was equivalent to
that of the center camera as if the vehicle had been shifted left or
right in the driving lane. In addition, all the camera images were
viewpoint transformed over a range so that data was created
corresponding to arbitrary shifts. These shifted images were paired
with steering commands that would guide the vehicle back to the center
of the lane.

The viewpoint transforms were done in the following way: first, the
assumption was made that the world is flat and therefore a pixel's
vertical index corresponded to the distance of the source from the
camera. The horizon was identified and all the pixels above the
horizon were considered to be at infinite distance and were left
untouched by the transform. The portion of the image below the horizon
was skewed in proportion to the distance in image space from the
horizon.

Of course, the world is not flat. Any objects above the real-world
ground plane were distorted by this method. Nevertheless, the
effectiveness of augmentation was proven by experiment.

It took a year from when our NVIDIA Holmdel group was formed to when
we had use of a drive-by-wire car. We initially tested our ideas using
a Husky~\cite{husky} mobile robot. Data was first gathered using the
Focus. With this data we trained a \gls{dnn} to produce a steering
angle given an image of the road. The trained network was then
installed on an NVIDIA DRIVE\texttrademark~PX system mounted on the
Husky. The Husky was skid steered, so the team had to empirically
translate the \gls{dnn} steering command to appropriate differential
wheel rotation commands to steer the Husky.

\begin{figure}[htb]
  \hfil
  \includegraphics[width=0.6\textwidth]{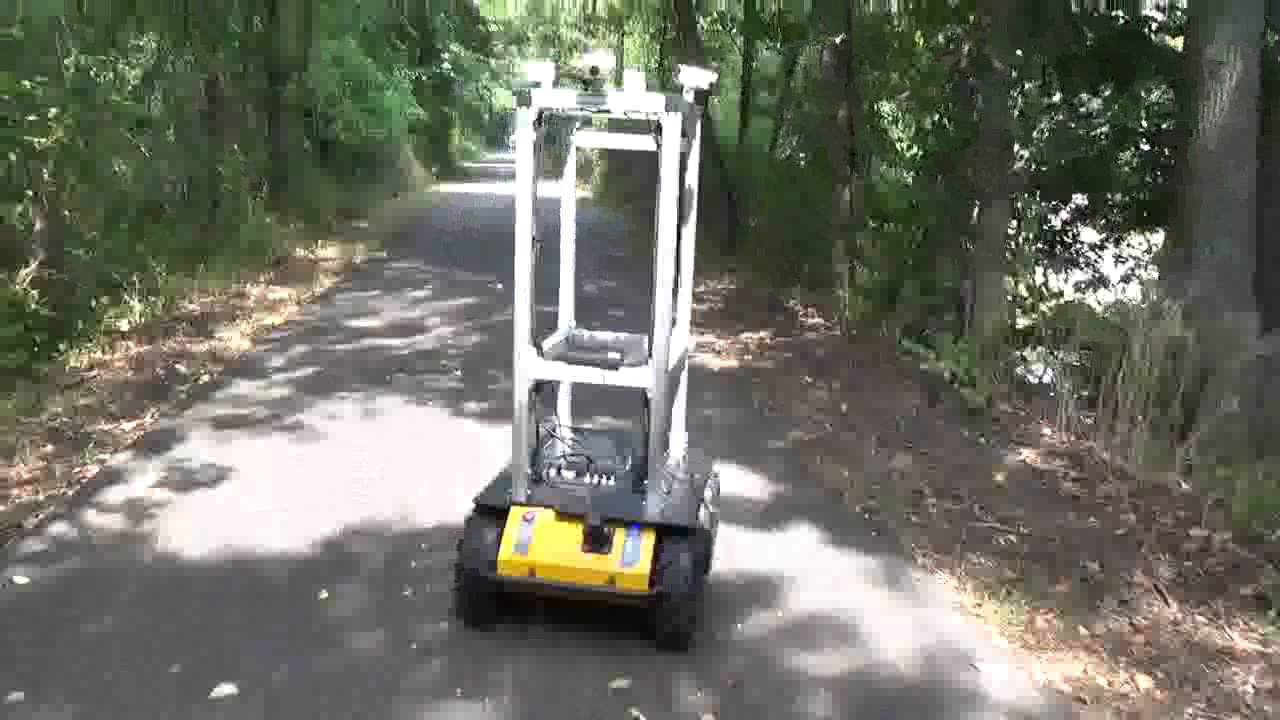}
  \caption{First field test with the Husky robot on a bicycle
    trail. Four cameras were mounted at about the same height where a
    windshield camera in a sedan would be. Only the front-facing
    camera was used in this experiment.}
  \label{fig-husky}
\end{figure}

The Husky could not be driven on public roads, but fortunately near
the NVIDIA office there was an off-road area with a defined path where
testing was possible (Figure~\ref{fig-husky}). Initial results were
disappointing --- the Husky would drive on the path for a few tens of
meters and then drift off the path. We suspected that the network,
using the perturbations, learned how to recover from shifts, but that
it could not recover from rotations in which the vehicle yaw axis is
misaligned with the direction of the road. To test this idea we added
image transforms to the training data corresponding to rotations
paired with the appropriate steering corrections. These transforms
were simple, all the pixels were shifted by the same amount either
left or right. The additional augmentation did the trick and the Husky
could drive on the bicycle path.

In February of 2016 we took delivery of a Lincoln MKZ that was
modified by the company AutonomouStuff for drive-by-wire operation and
equipped with cabling for all the sensors. Also that month we moved to
BellWorks, the new name for the former Bell Labs building in Holmdel
where much of today's machine learning theory and technology was
created. The BellWorks complex features a garage for the MKZ as well
as a few miles of private roads which are ideal for close-course
testing of a self-driving car.

During the next month we installed cameras and the NVIDIA
DRIVE\texttrademark~PX board along with some support electronics on
the MKZ. We created a rudimentary self-driving system by coupling
PilotNet with the MKZ's \gls{acc} for speed control. To our delight
the car drove well on the BellWorks internal roads. (Of course, as in
all our tests, we observed all relevant rules and regulations,
including having a safety driver who was always on the alert to take
full control of the car if needed.) The next day we took the car onto
Middletown Road, adjacent to BellWorks. This is a twisty, hilly road
that runs for a few miles and at the time was poorly paved. Again, the
car drove credibly. The next day the self-driving MKZ (with a few
interventions) took team members to a nearby restaurant to
celebrate. The system was first shown to the public at the \gls{gtc}
San Jose in the Spring of 2016
(\href{https://youtu.be/qhUvQiKec2U}{available on Youtube}) and later
at \gls{gtc} Europe in the Fall of 2016
(\href{https://www.youtube.com/watch?v=-96BEoXJMs0}{available
  on Youtube}).

\bigskip
Over the next nine months the self-steering system, now known as
PilotNet, was rapidly improved, culminating in a public demonstration
on a closed course at the 2017 \gls{ces} in Las Vegas.

\begin{figure}[htb]
  \hfil
  \includegraphics[width=\textwidth]{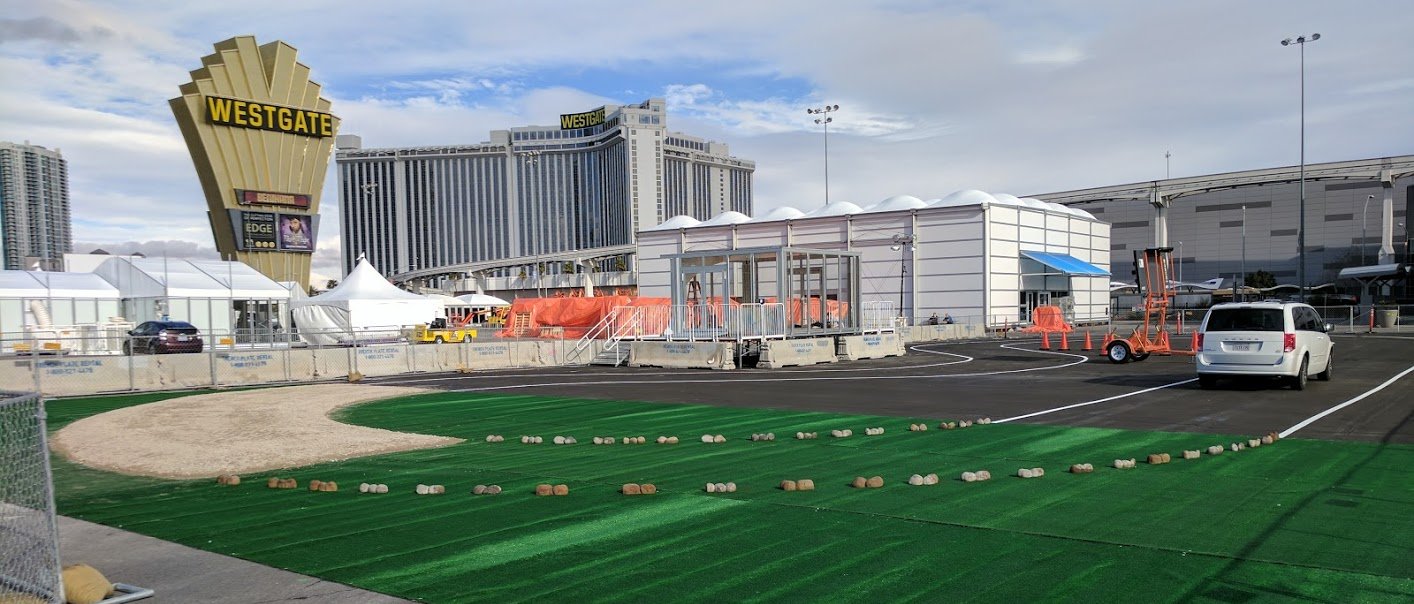}
  \caption{The \gls{ces} 2017 demo course before the show opened.}
  \label{fig-ces}
\end{figure}

The \gls{ces} demo was set up just before Christmas in an unusual
rainy spell in Las Vegas. The car was tuned to drive well in the demo
by using training data recorded on the demo course which included a
variety of road surfaces (paved, gravel, grass) and some movable
construction obstacles. When the weather cleared and the sun came out,
the car's performance declined. Also, as the start of CES approached,
the fencing and banners around the test course were changed and the
car failed to stay in the lanes. Rapid on-site retraining of PilotNet
with additional data that included different lighting and different
objects at the track perimeter fixed the problems and the car drove
well throughout the event. During \gls{ces}, hundreds of passengers
rode in the back seats of the PilotNet driven MKZ and an SUV provided
by Audi. The 2017 \gls{ces} demo course is shown in
Figure~\ref{fig-ces} and reported by various on-line publications, for
example by \href{https://www.youtube.com/watch?v=FABftuXUOxE}{The
Verge}.

We defined a metric for benchmarking PilotNet performance: \gls{mdbf}
which is the average distance traveled between required human
interventions. Early versions of PilotNet were trained on a few hours
of human driving data on both highways and local roads. Early PilotNet
had an \gls{mdbf} of about 10\km: it was able to steer an average of
about 10\km\ in the center lane of the Garden State Parkway near
Holmdel, NJ. Here, there are three lanes in each direction of the
limited access highway. PilotNet was also able to steer on local
roads, but a human operator had to intervene at intersections since
early PilotNet had no route planning capability and could not initiate
turns.

\subsection{Learning a Trajectory}
\label{sec-traj}
Early PilotNet produced a steering angle as output. While having the
virtue of simplicity, this approach has some drawbacks. First, the
steering angle does not uniquely determine the path followed by a real
car. The actual path depends on factors such as the car dynamics, the
car geometry, the road surface conditions, and the transverse slope of
the road (banks or domes). Ignoring these factors can prevent the
vehicle from staying centered in its lane. In addition, just producing
the current steering angle does not provide information about the
likely course (intent) of the vehicle. Finally, a system that just
produces steering is hard to integrate with an obstacle detection
system.

To overcome these limitations, PilotNet now outputs a desired
trajectory in a 3D coordinate frame relative to the car coordinate
system. An independent controller then guides the vehicle along the
trajectory. Following the trajectory, rather than just following a
steering angle, allows PilotNet to have consistent behavior regardless
of vehicle characteristics.

An added bonus in using a trajectory as an output is that it
facilitates fusing the output of PilotNet with information from other
sources such as maps or separate obstacle detection systems.

To train this newer PilotNet, it is necessary to create the desired
target trajectory. Initially, we used the trajectory followed by a
human driver in the data collection car as the target trajectory. This
trajectory was extracted from the vehicle pose derived from recorded
vehicle speed, \gls{imu}, and \gls{gps} sensor data. We later
discovered that the human-driven trajectory, coupled with data
augmentation, resulted in poor on-the-road driving (see
Section~\ref{sec-tuning}). We now use the lane centerline, as
determined by human labelers, as the ground-truth desired
trajectory. See Section~\ref{sec-labeling}.

\subsection{Additional PilotNet Outputs}
PilotNet now provides multiple outputs that correspond to seven
possible trajectories. An external navigation system chooses among
these trajectories to guide the vehicle. The possible trajectories
are:
\begin{compactitems}
  \item Lane stable (keep driving in the same lane) 
  \item Change to left lane (first half of maneuver)
  \item Change to left lane (second half of maneuver)
  \item Change to right lane (first half of maneuver)
  \item Change to right lane (second half of maneuver)
  \item Split right (\eg\ take an exit ramp)
  \item Split left (\eg\ left branch of a fork in the road)
\end{compactitems}

In addition, PilotNet now predicts lane boundaries that are inferred
even when there are no lines painted on the road. The lane boundary
output allows third parties to plan their own trajectories using the
path perception provided by PilotNet.

\section{Data Collection and Curation}
\label{sec-collection}
Creating a huge, varied corpus of clean, accessible data is one of the
most critical and resource-intensive aspects encountered in creating a
high-performance, learned self-driving system. In this section we
describe our data collection and curation process.

\subsection{Early Data Collection}
To see if our ideas held promise, we did some rapid prototyping. As
previously mentioned, we trained early versions of PilotNet using
video from GoPro cameras mounted with suction cups on our data
collection cars. See Figure~\ref{fig-focus1}. The cars were driven
primarily within Monmouth County, New Jersey under various weather and
lighting conditions. Data was recorded using an NVIDIA
DRIVE\texttrademark~PX system.

\subsection{Hyperion Data}
The early data was adequate for demonstrating a proof-of-concept, but
to get to a high-performance product, a massively scaled data
collection process was required. Recognizing this need, in 2017 NVIDIA
embarked on a large-scale data collection campaign. This effort, known
as Hyperion~\cite{hyperion7}, included building a fleet of Ford Fusion and Ford
Mondeo cars to collect data not only for PilotNet but also for
numerous other subsystems required for a complete driving solution.

\begin{figure}[htb]
  \hfil
  \includegraphics[width=0.6\textwidth]{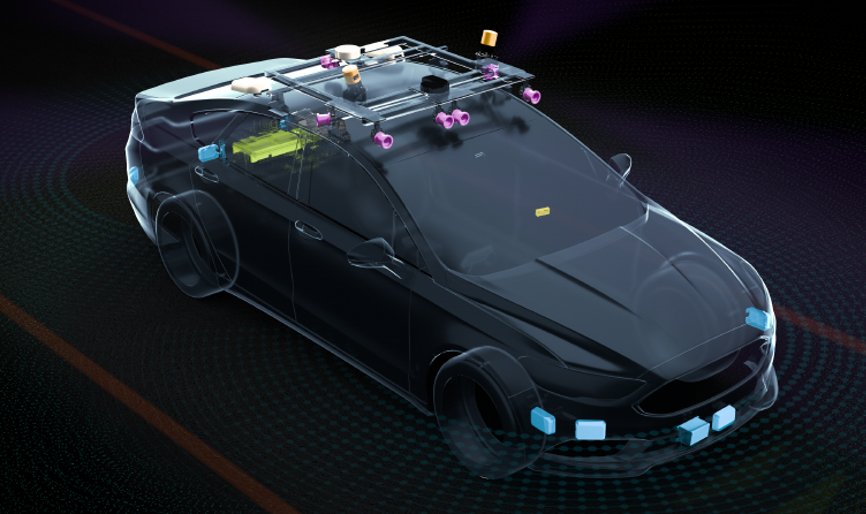}
  \caption{NVIDIA Hyperion data collection vehicle sensor
    layout~\cite{hyperion7}.}
  \label{fig-h7}
\end{figure}

Collected data included imagery from 12 automotive-grade cameras,
returns from 3 lidars and 8 radars, \gls{gps} data, as well as
\gls{can} data such as steering angle and speed. Computing for the
data collection platform was provided by two NVIDIA DRIVE™ PX 2
systems. Data was collected in the U.S., Japan, and Europe with routes
chosen to maximize diversity. Figure~\ref{fig-h7} shows one of the
Hyperion data collection cars.

\section{The Training Pipeline}
\label{sec-training}
Training PilotNet is a complex, multi-stage process that requires
manipulating huge quantities of data as well as extensive
computing. To train PilotNet efficiently we developed a comprehensive
automated data pipeline. The pipeline lets us simultaneously explore
different adjustments to network architecture, data selection, and
image processing. In addition, the pipeline allows us to run
experiments reproducibly; much effort was expended to ensure
deterministic behavior. Elements of the pipeline are described below.

\subsection{The Labeling Process}
\label{sec-labeling}
\begin{figure}[htb]
  \hfil
  \includegraphics[width=0.75\textwidth]{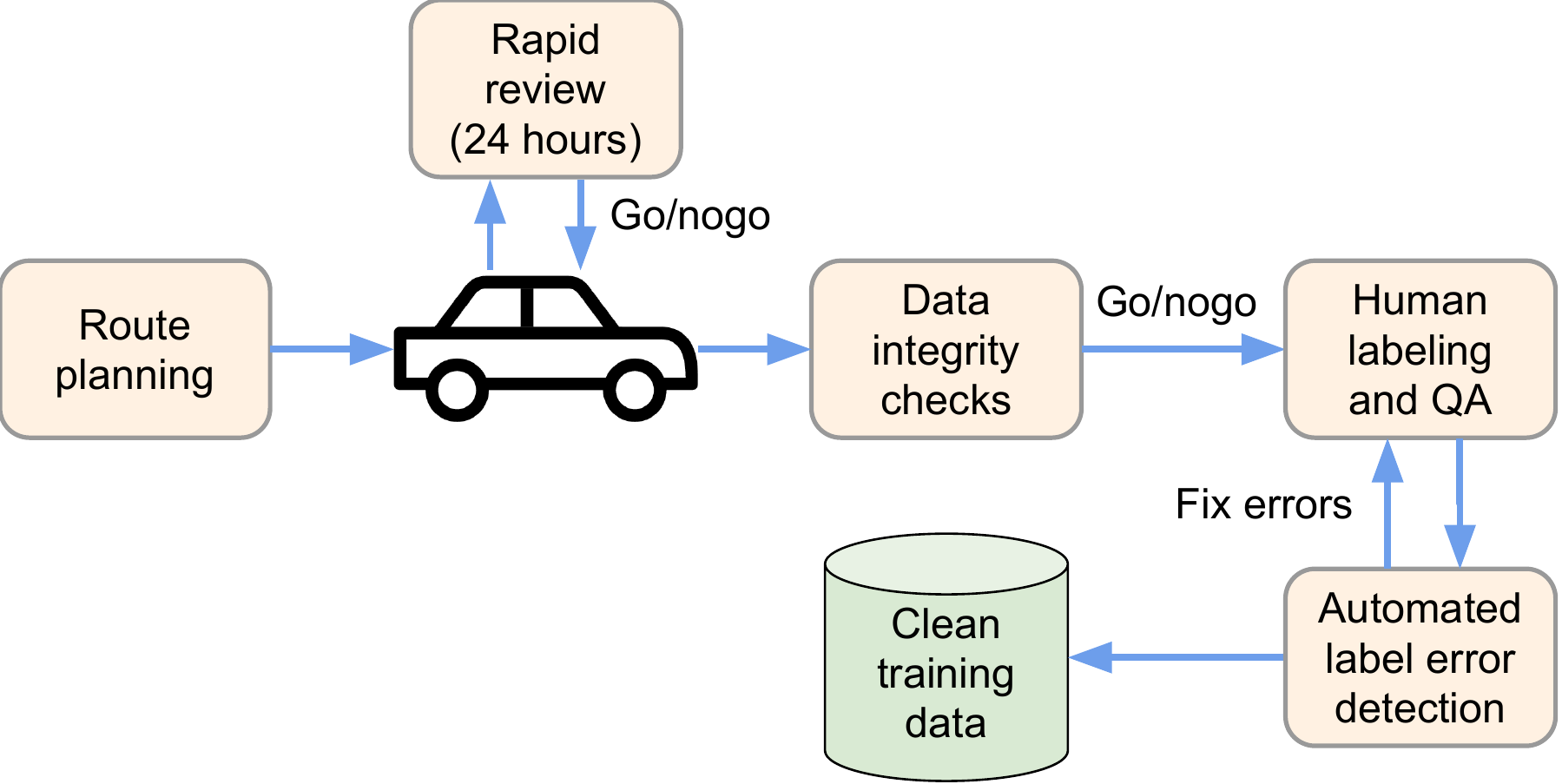}
  \caption{Diagram outlining Hyperion data collection, rapid field
    review, labeling, \gls{qa}, and data archiving. The process
    strives to produce and store only high-quality data. Rapid field
    review ensures that we can react to basic data quality problems
    quickly (\eg, if sensors are not recording). The data integrity
    checks run in the cloud and analyze the data quality in detail
    (\eg, proper time stamps). However, it can take several weeks
    before the data becomes available in the cloud.}
  \label{fig-labeling-proc}
\end{figure}

Collected data was processed using the PilotNet labeling tool in which
human labelers can quickly examine recorded video and segment it into
clips representing different driving maneuvers (lane stable, lane
change, lane split, merge, turn). In addition, the data is tagged for
different conditions such as road type (multi-lane highway,
single-lane highway, unmarked road), road condition (dry, wet) and for
special events such as avoiding an obstacle. See
Figure~\ref{fig-labeling-proc}.

The labels go through a \gls{qa} process, where multiple labelers
check each other's work. The best labelers become quality assurance
staff, checking on the work of the other labelers. In addition, we run
a trained PilotNet network over the new labeled data and flag
instances where the network trajectory output has a large discrepancy
compared to the label. These segments are then checked by the most
experienced labelers to make sure the labels are correct. Bad segments
are then re-labeled. We refer to this process as ``bad appling,'' since
we discovered that a few mislabeled critical segments can create
unexpected failure modes in the system.

As data is cleaned, labeled, and archived, small chunks of this new
data are added to our active training data set and a preliminary new
PilotNet is trained. Test results are compared with the previous
version of PilotNet to make sure that that additional data improves
overall performance. If performance degrades, the new data undergoes
further scrutiny to uncover the reason for the degradation.

\begin{figure}[htb]
  \hfil
  \includegraphics[width=\textwidth]{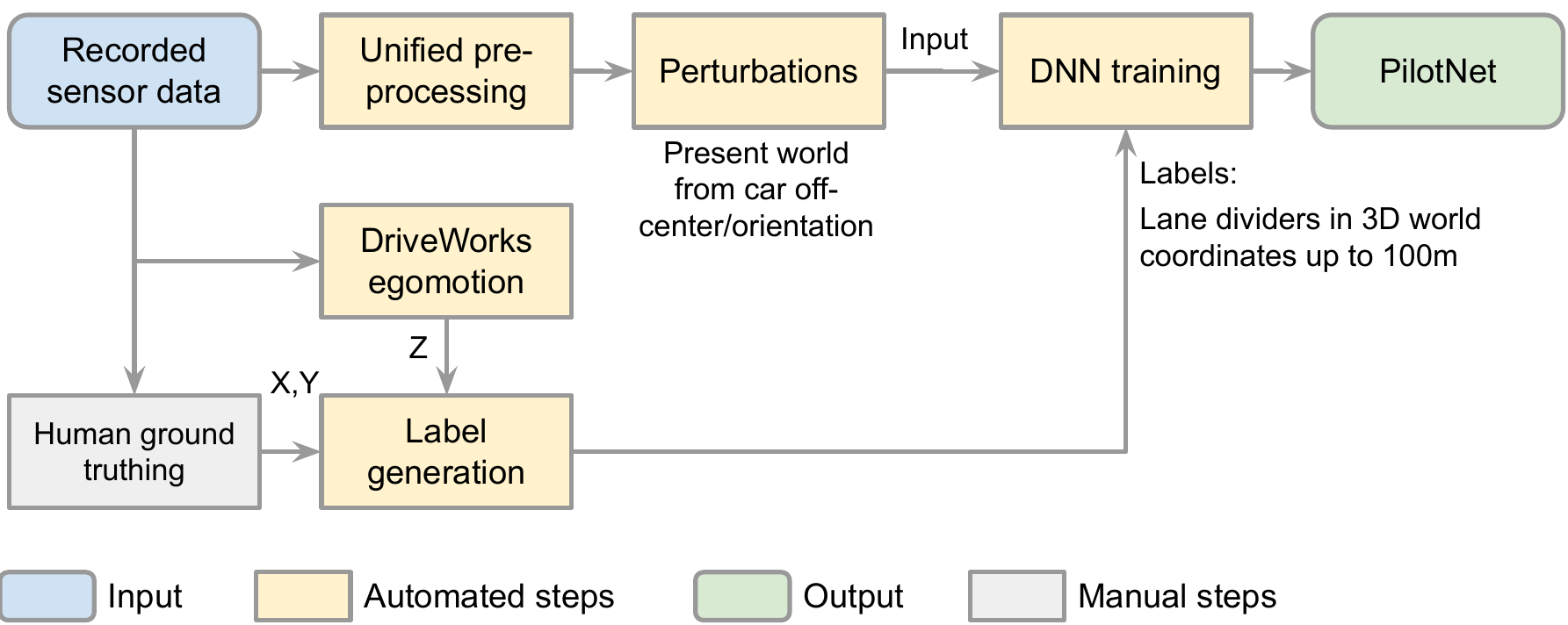}
  \caption{A data-flow diagram showing how labels are generated for
    training a PilotNet network. The input on the top left and the
    manual steps on the bottom left correspond to processes outlined
    in Figure~\ref{fig-labeling-proc}.}
  \label{fig-label-flow}
\end{figure}

PilotNet employs a supervised learning framework to train a neural
network to predict trajectories that the vehicle should follow. To
train such a network, labelers mark the edges of the lanes driven by
professional human drivers. The points along the center line between
these edges are defined as the desired $X$ and $Y$ trajectories and
become the ground truth $X$ and $Y$ labels. The training trajectory's
vertical ($Z$) component is calculated using the vehicle egomotion,
\ie, the relative position and orientation of the vehicle as a
function of time. The egomotion module, which was created by the
NVIDIA DriveWorks~\cite{nvdrive} team, uses IMU, odometry, and GPS
recordings with a vehicle motion model to deduce the egomotion of the
vehicle in 3D. The data-flow diagram for training PilotNet is shown in
Figure~\ref{fig-label-flow}.

During inference time, PilotNet provides desired trajectories in the
vehicle's local coordinate frame. In a preprocessing step, PilotNet's
trajectory generation module first creates the trajectory of the
entire recording in a global coordinate frame. Then a library of
trajectory transformation algorithms extracts segments of this
trajectory data that corresponds to each training frame. The library
also transforms the trajectory segment from the global world
coordinate frame to the vehicle's local coordinate frame that is used
during training.

\subsubsection{Training Label Generation}
Human labelers annotate driving maneuvers (lane stable, lane change,
lane split, merge, turn), road types (multi-lane highway, single-lane
highway, unmarked road), road conditions (dry, wet) and special events
such as obstacle evasion and speed changes to comply with traffic
rules (\ie, stopping at a traffic light). Some of these annotations
are used for generating control inputs sent to the network at training
time. The rest are used for filtering and sampling the data.

The trajectory label, derived as described in Section~\ref{sec-traj},
consists of 100 3D points spaced one meter apart. These points are
referenced relative to the car coordinate system.

\subsection{Preprocessing}
Camera image characteristics vary depending on camera type and on
camera location and orientation. Preprocessing reformats image
training data, compensating for these variations. The remainder of
this section describes this preprocessing pipeline.

\subsubsection{Image Pinhole Rectification} First, camera images are
reformatted so that they appear as if they were recorded by an ideal
(pinhole) camera. This step removes distortions that are particular to
the camera lens, rendering the image lens-independent. Thus PilotNet
can be made robust to variations in intrinsic camera parameters. This
means the network can be trained on images gathered from one set of
cameras and later produce driving trajectories for images created by a
different set of cameras.

\subsubsection{Viewpoint Transformation}
\label{sec-viewpoint-transform}
Second, camera images are transformed so that they appear to be
captured from a standard position and orientation on the vehicle. All
points in the training trajectory are measured relative to the rear
axle of the vehicle, so the transformed images appear as if the camera
is 1.47\m\ above the rear axle and 1.77\m\ in front of the rear axle
along the center line of the car. These numbers come from the actual
camera placement on most of our data collection cars. However, we want
PilotNet to work on other cars in which the camera may be in a
different location. With this ``viewpoint'' transformation the
processed images become nearly independent of the precise camera
placement on the vehicle.

We note that the viewpoint transformation cannot be perfect, since
there are bits of the road ahead that may be visible from one camera
location and not visible from another. However, these discrepancies
are small when we consider portions of the road that are beyond a few
meters in front of the vehicle.

\begin{figure}[htb]
  \hfil
  \begin{subfigure}{0.45\textwidth}
    \includegraphics[width=1.0\linewidth]{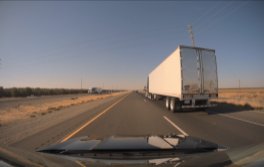}
    \caption{Original camera image}
  \end{subfigure}
  \hspace{1ex}
  \begin{subfigure}{0.45\textwidth} 
   \includegraphics[width=1.0\linewidth]{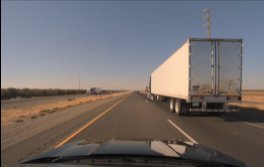}
    \caption{Pinhole-transformed image}
  \end{subfigure}

  \medskip
  \hfil
  \begin{subfigure}{0.45\textwidth}
    \includegraphics[width=1.0\linewidth]{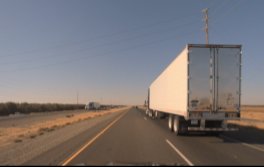}
    \caption{Viewpoint-transformed to standard camera position}
  \end{subfigure}
  \hspace{1ex}
  \begin{subfigure}{0.45\textwidth}
    \includegraphics[width=1.0\linewidth]{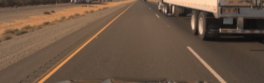}
    \caption{\gls{roi} cropped image}
  \end{subfigure}
  \caption{Examples of rectifying and transforming an image to a
    standard format and resulting \gls{roi}.}
  \label{fig-label-roi-crop}
\end{figure}

Third, images are cropped to a \gls{roi}. The boundaries of this ROI
are in 3D world coordinates rather than image space. The specifics of
the ROI are chosen in the following way: first we set the horizontal
field of view to be 53\degr\ wide. Next we assume the ground is flat and we
choose the top of the ROI to align with to the horizon. Finally we
adjust the bottom of the ROI to correspond to a section of the ground
that is 7.6\m wide. With these adjustments in mind, the images are
linearly scaled so that the resulting image is 209 pixels wide and 65
pixels high. Refer to Figure~\ref{fig-label-roi-crop}.

By choosing these parameters we eliminate the sky which has little
bearing on driving. Provided the camera has sufficient resolution, we
create standardized images that are largely independent of camera
properties.

\subsection{Training Data Augmentation}
\label{sec-training-data-aug}
We observed in early experiments that, when PilotNet's neural network
was trained only with samples where the vehicle is aligned with the
target trajectory, the network had challenges predicting the correct
trajectory if the vehicle deviated from the center of the road. This
occurred because off-road-center driving is outside of the original
training data. As a result, the driving system could not recover from
a series of network errors, controller errors, or environmental
factors that would cause the vehicle to deviate from the lane
center. This is a well known problem with imitation
learning~\cite{Pom89,pmlr-v9-ross10a,NIPS2019_9343}.

\begin{figure}[htb]
  \begin{subfigure}{0.32\textwidth}
    \includegraphics[width=1.0\linewidth]{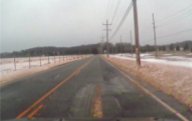}
    \caption{Original image}
  \end{subfigure}
  \hfill
  \begin{subfigure}{0.32\textwidth} 
   \includegraphics[width=1.0\linewidth]{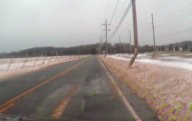}
    \caption{1\m\ shift right}
  \end{subfigure}
  \hfill
  \begin{subfigure}{0.32\textwidth} 
    \includegraphics[width=1.0\linewidth]{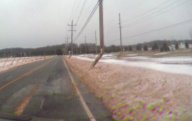}
    \caption{1\m\ shift right, 10\degr\ rotate right}
  \end{subfigure}

  \medskip
  \begin{subfigure}{0.32\textwidth}
    \hfil
    \includegraphics[width=0.45\linewidth]{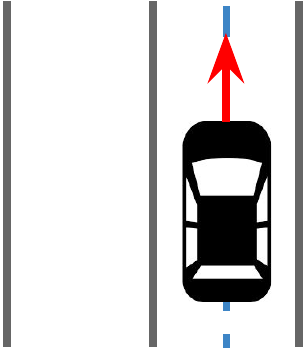}
  \end{subfigure}
  \hfill
  \begin{subfigure}{0.32\textwidth}
    \hfil
    \includegraphics[width=0.45\linewidth]{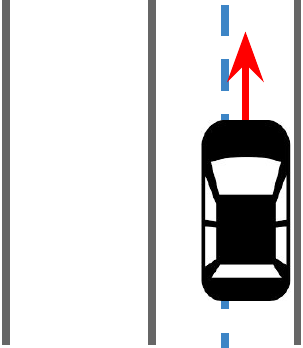}
  \end{subfigure}
  \hfill
  \begin{subfigure}{0.32\textwidth}
    \hfil
    \includegraphics[width=0.45\linewidth]{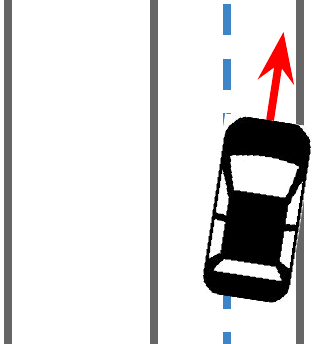}
  \end{subfigure}
  \caption{Illustration of transforming camera data with a flat-world
    assumption to emulate a shift and rotation of the vehicle.}
  \label{fig-shift-rot}
\end{figure}

To mitigate this issue, we augment our data with training images that
make the vehicle appear shifted from the lane center and/or rotated
from the lane direction. We use the same viewpoint transforms we use
to transform camera images to a nominal position to emulate a shift
and rotation of the camera. Figure~\ref{fig-shift-rot} illustrates
the viewpoint transformations used to augment training data.

\begin{figure}[htb]
  \hfil
  \includegraphics[width=0.4\textwidth]{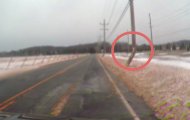}
  \caption{Resulting artifacts that occur because of an incorrect
    flat-world assumption. Vertical features whose base is below the
    horizon in the image are distorted. As an example, look at the
    utility pole marked by the red circle.}
  \label{fig-artifacts}
\end{figure}

There is one caveat of using viewpoint transforms for virtually moving
a camera: we do not have accurate 3D geometric information about the
world, so we assume a flat world and transform the image according to
that geometry. This creates distortion artifacts on any object in the
image that does not fit our flat-world assumption. See
Figure~\ref{fig-artifacts}.

Because these artifacts carry information about the shift and rotation
of the augmentation, it is possible that these artifacts could become
a dominant signal that the network uses to learn the training labels
rather than using the image of the road. A more detailed discussion of
this effect is presented in Section~\ref{sec-tuning}. To lessen this
possibility, we record from multiple (typically three) cameras placed
at different shifts on the vehicle and stochastically select which of
these cameras to use in a training example. Because the cameras each
have a different original position, the resulting distortion artifacts
are different depending on which camera we are using for the training
example. This procedure introduces variation to the distortion signal
to discourage the network from learning a correlation between the
distortion artifacts and the off-center position of the vehicle. In
addition, human driving behavior during data collection frequently
deviates from the center of the lane (standard deviation typically is
20\cm). This adds additional shift and rotation variations relative
to the true center trajectory we use as the training label today.

\section{PilotNet Neural Network Architecture}
\label{sec-nnarch}
\begin{figure}[htb]
  \begin{subfigure}{0.19\textwidth}
    \hfil
    \includegraphics[width=0.6\linewidth]{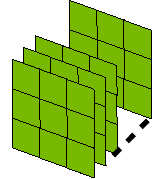}
    \caption{2D convolution}
  \end{subfigure}
  \hfill
  \begin{subfigure}{0.19\textwidth}
    \hfil
    \includegraphics[width=0.6\linewidth]{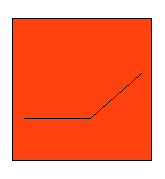}
    \caption{ReLU}
  \end{subfigure}
  \hfill
  \begin{subfigure}{0.19\textwidth}
    \hfil
    \includegraphics[width=0.6\linewidth]{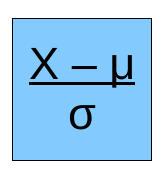}
    \caption{2D Batch Norm}
  \end{subfigure}
  \hfill
  \begin{subfigure}{0.19\textwidth}
    \hfil
    \includegraphics[width=0.6\linewidth]{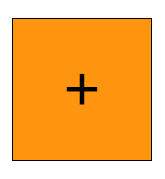}
    \caption{Linear Addition}
  \end{subfigure}
  \hfill
  \begin{subfigure}{0.19\textwidth}
    \hfil
    \includegraphics[width=0.6\linewidth]{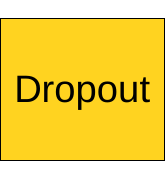}
    \caption{Spatial Dropout}
  \end{subfigure}
  \caption{The five basic building blocks of PilotNet shown in
    Figure~\ref{fig-basic-res}}
  \label{fig-nn-blocks}
\end{figure}

\begin{figure}[htb]
  \hfil
  \includegraphics[width=\textwidth]{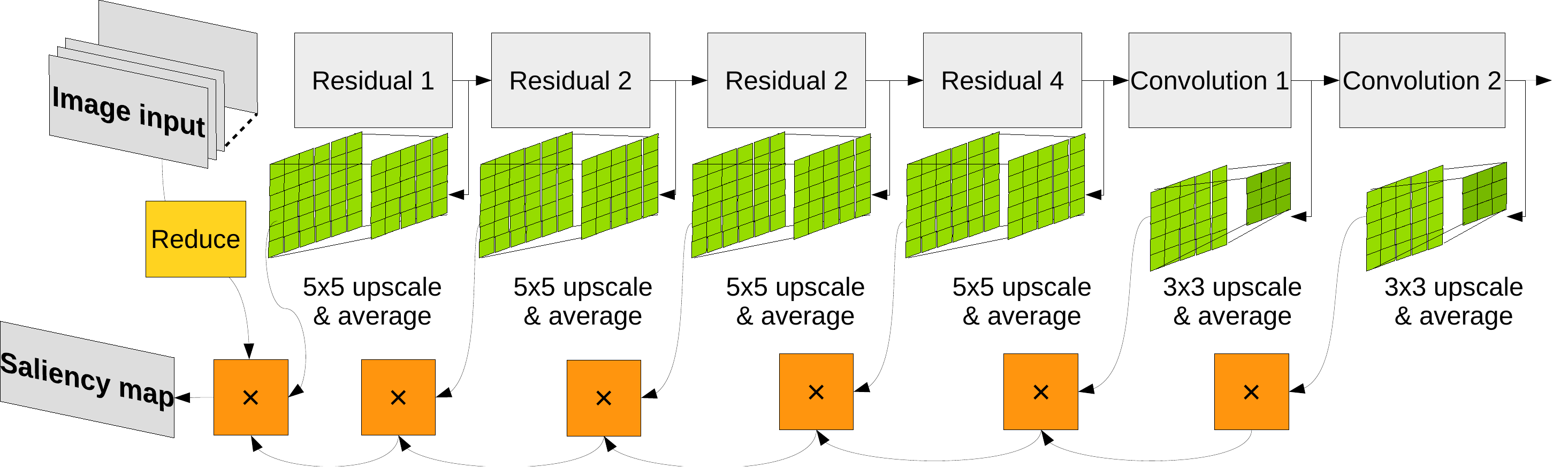}
  \caption{Outputs of the residual and convolution layers are
    upscaled, averaged, and multiplied to create a single-channel
    saliency map (\ie, finding the pixels that are most crucial in
    determining the driving trajectory~\cite{explain-e2e-2017}.}
  \label{fig-pn-outputs}
\end{figure}

Early PilotNet followed a typical convolutional neural network
structure~\cite{lecun-89e}. More recent versions feature a modified
ResNet structure~\cite{1512.03385}. See
Figures~\ref{fig-nn-blocks}-–\ref{fig-pn-outputs}.

\begin{figure}[htb]
  \hfil
  \includegraphics[width=\textwidth]{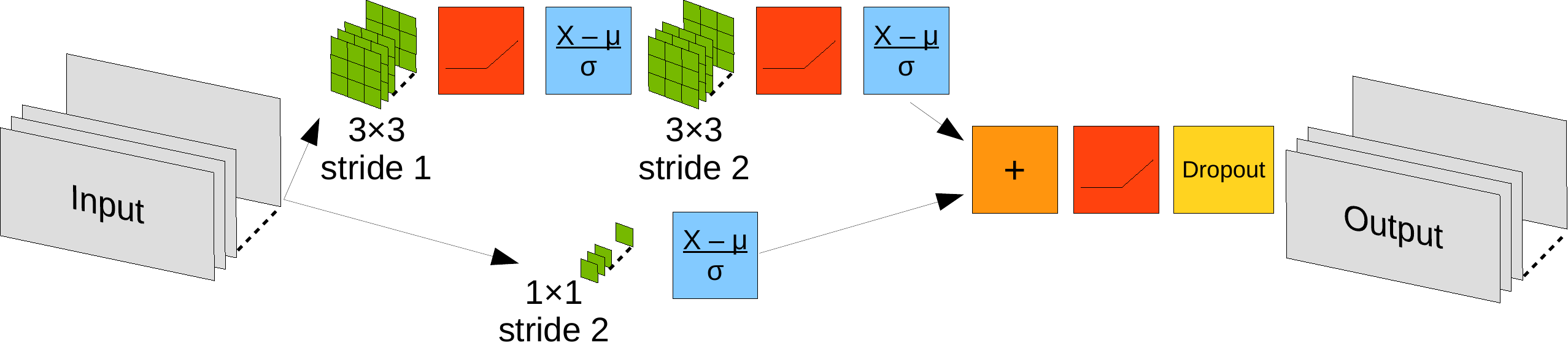}
  \caption{A basic residual block. All 3\x3 kernels use padding
    of~1. Padding adds $N$ pixels at the edges of the input matrix,
    which is necessary for the convolution operator to be properly
    applied to the cells at the edges.}
  \label{fig-basic-res}
\end{figure}

\begin{figure}[htb]
  \hfil
  \includegraphics[width=\textwidth]{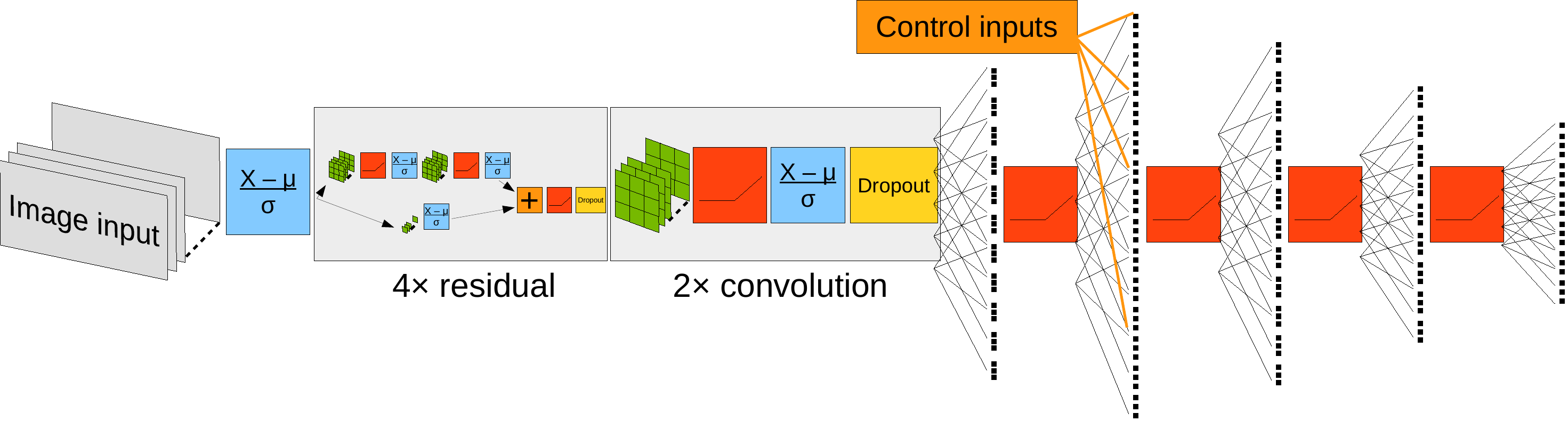}
  \caption{The full PilotNet training architecture. Sizes of the
    intermediate steps appear in Table~\ref{tab-pilotnet}. Control
    inputs to control network driving behavior (split, lane change,
    etc.) are added to the inputs of the second fully connected
    layer.}
  \label{fig-nn-pilotnet}
\end{figure}

\begin{figure}[htb]
  \hfil
  \includegraphics[width=.8\textwidth]{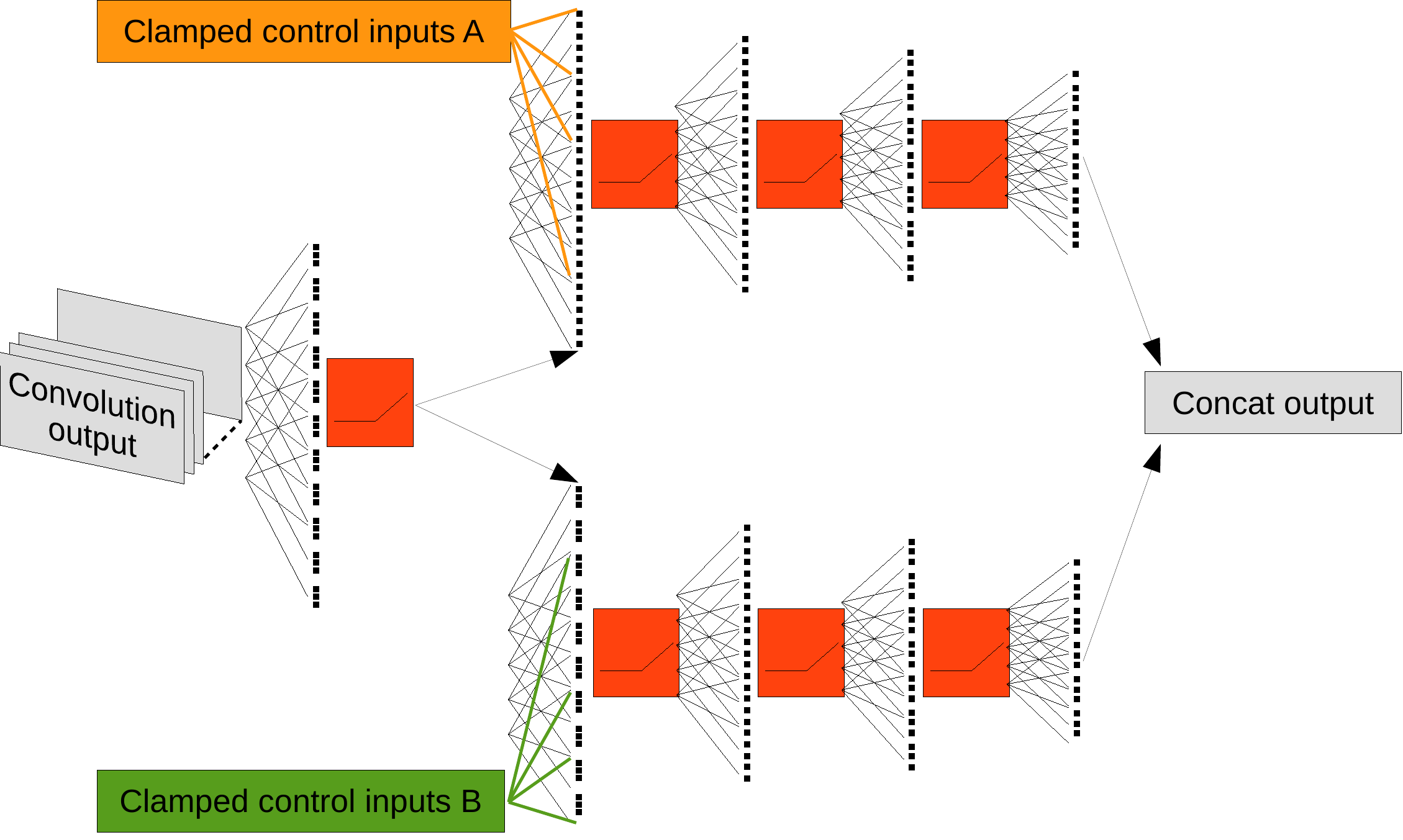}
  \caption{The exported version of the PilotNet network with
    modifications for simultaneous trajectory outputs without
    requiring control input selection. The fully connected layers are
    cloned and the control inputs are frozen for each output. This can
    be done with any number of behaviors, only two are visualized
    here. For run-time optimization it is also possible to convert the
    cloned linear layers into sparse convolutions.}
  \label{fig-pn-exported}
\end{figure}

\begin{table}
  \hfil
  \tabulinesep=0.5ex
  \begin{tabu}{|c|c|}
    \hline
    {\bf Layer name}    & {\bf Output size}     \\
    \hline\hline
    Batch norm 1        & 3\x113\x209            \\
    \hline
    Residual 1          & 48\x57\x105           \\
    \hline
    Residual 2          & 72\x30\x53            \\
    \hline
    Residual 3          & 96\x15\x27            \\
    \hline
    Residual 4          & 128\x8\x14            \\
    \hline
    Convolution 1       & 192\x6\x12            \\
    \hline
    Convolution 2       & 256\x4\x10            \\
    \hline
    Flatten             & 10,240                \\
    \hline
    Linear 1            & 256                   \\
    \hline
    Linear 2            & 256                   \\
    \hline
    Linear 3            & 128                   \\
    \hline
    Linear 4            & 96                    \\
    \hline
    Linear 5            & Target                \\
    \hline
  \end{tabu}
  \caption{Output sizes of PilotNet layers. The exact number of
    outputs is determined by the training target (number of paths,
    number of dimensions, etc.).}
  \label{tab-pilotnet}
\end{table}

\subsection{Control Inputs}
To proceed beyond simple lane following, an autonomous car must be
able to switch lanes, negotiate lane merges and splits, and execute
turns. PilotNet has been extended to support these tasks. As shown in
Figure~\ref{fig-nn-pilotnet}, we can accommodate an arbitrary number
of behaviors by creating a control input vector for each behavior and
feeding it into a clone of the last four linear layers. A control system
receives all the outputs of the network and decides which to use based on
the current maneuver. For runtime execution (inference), the trained
network is exported using NVIDIA TensorRT, which generates optimized
CUDA kernel schedules to increase inference performance on the NVIDIA
DRIVE\texttrademark~AGX  in the vehicle.

\subsection{Large-Scale Training}
PilotNet currently trains on millions of frames. All data manipulation
is executed in parallel on one of NVIDIA's internal GPU compute
clusters. Training a PilotNet network from scratch to peak driving
performance takes about two weeks.

Instead of augmenting our training data online, we augment the samples
prior to the training step. By processing the data once, saving it to
disk, and reusing it in each epoch we save considerable compute
time. Training samples are shuffled during the off-line augmentation
process so that data can be read sequentially during training.

The training samples are considerably smaller in size compared to the
full-resolution video frames; therefore loading the preprocessed
samples from disk by itself, reduces the compute and I/O bandwidth
required per epoch. In addition, the augmented data can be re-used
across multiple experiments (such as neural architecture search,
hyperparameter tuning, or simply training with multiple seeds for the
random number generators), further reducing the compute requirements
of developing PilotNet.

\begin{figure}[htb]
  \hfil
  \includegraphics[width=\textwidth]{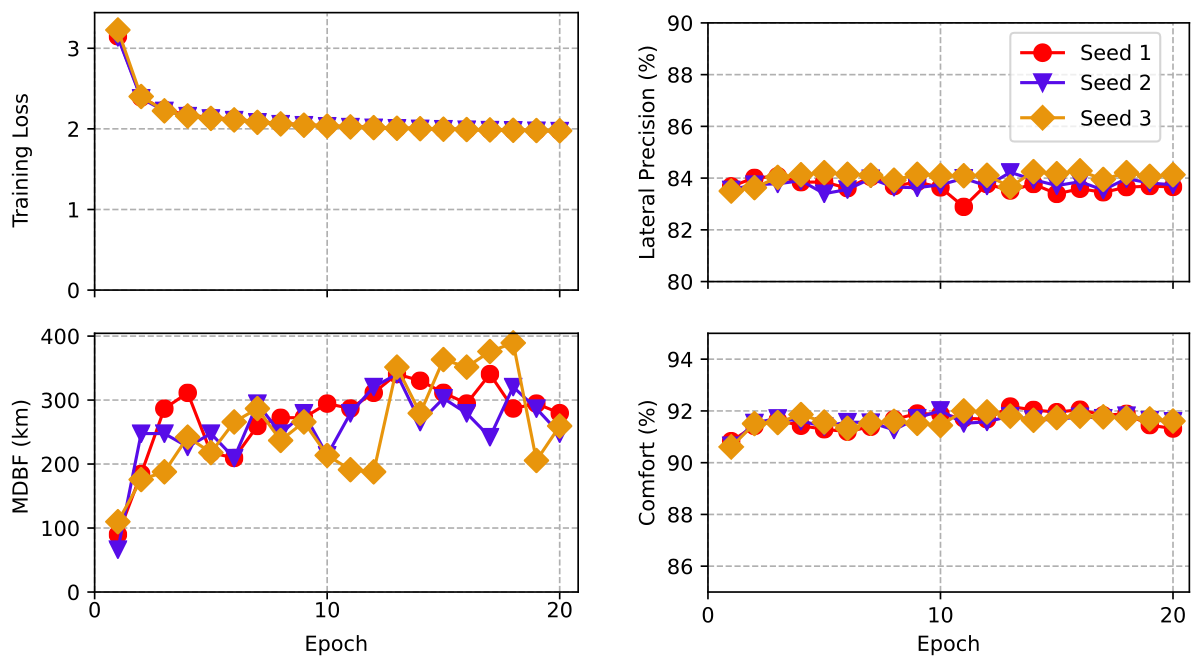}
  \caption{Evolution of training loss and driving performance during
    training, measured for three different neural networks initialized
    with different seeds. See Section~\ref{sec-nnarch} for the definition
    of the performance metrics.}
  \label{fig-training-loss}
\end{figure}

Processed and augmented frames, as well as the labels for these
frames, are stored as flat serialized binary files for fast data
I/O. We developed a custom data storage library in order to write and
access these files. The library is implemented in C++, and has Python
bindings for easy integration with the PyTorch data loader.

After each epoch of training, we launch a parallel task to measure the
performance of the network in augmented resimulation (see
Section~\ref{sec-aug-resim} below). This way, we not only monitor the
training loss, but also the metrics that are more relevant to the task,
\ie, the real-world driving performance. Figure~\ref{fig-training-loss}
shows the per-epoch training loss and driving performance for three
networks initialized with different seeds.

\subsection{Testing the Network: The Augmented Resimulator}
\label{sec-aug-resim}
We require a means to evaluate and compare different versions of
PilotNet. The most direct test is real-world on-road testing. However,
real-world tests are time consuming, not easily reproduced, and have
risk. Simulated test environments can help alleviate some of these
issues, but simulations may not be representative of the real
world. This is a particular concern for vision-based systems like
PilotNet, where road textures, glare, or even chromatic aberration
caused by different speeds of capturing the \gls{rgb} channels in the
camera can affect real-world driving. Producing a photo-realistic
simulation can be a challenge in itself.

\begin{figure}[htb]
  \hfil
  \includegraphics[width=\textwidth]{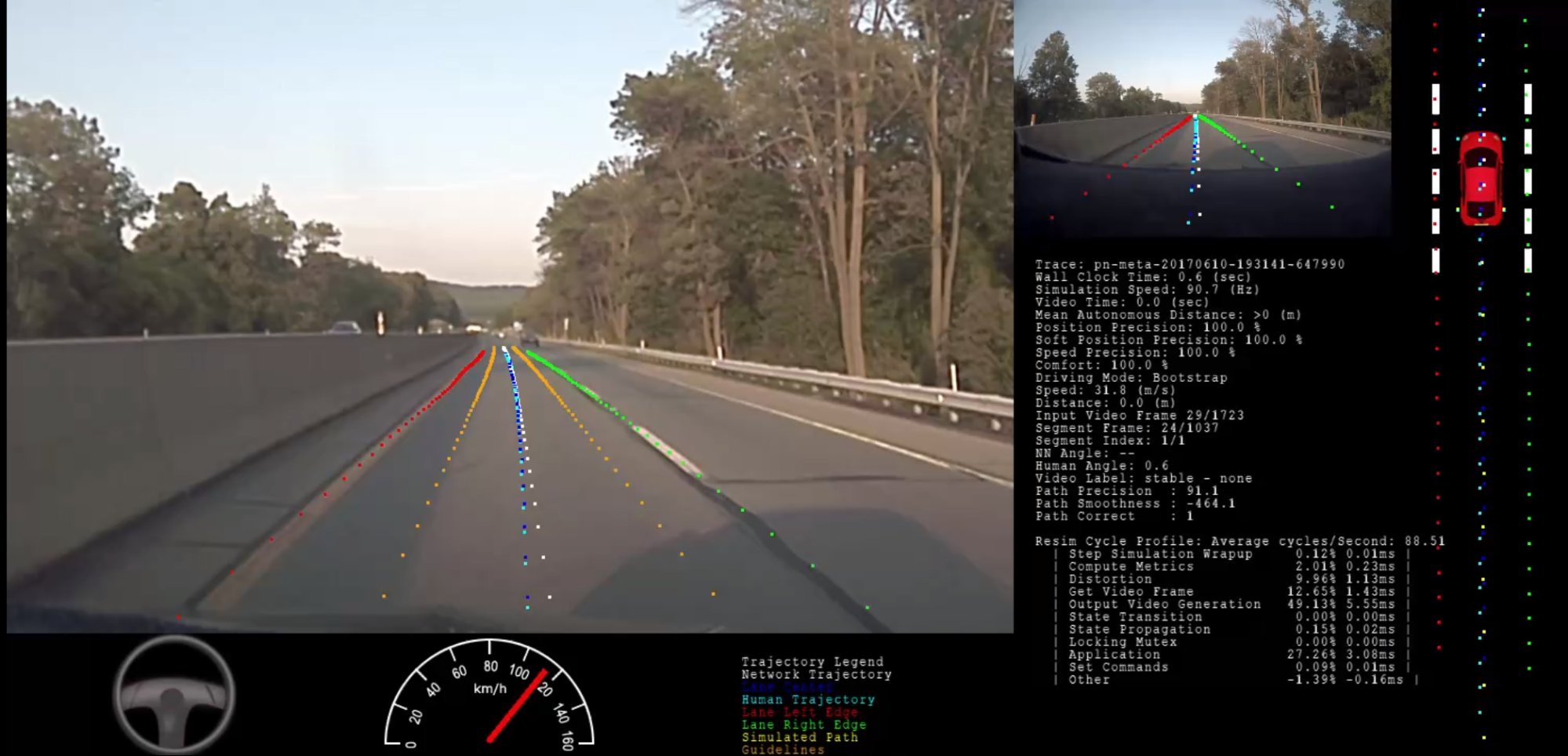}
  \caption{Screenshot of the Augmented Resimulator. The car icon in
    the upper right corner shows the position of the resimulated
    vehicle in its lane. The photographic image in the upper right is
    the original image captured while the data collection car was
    driven by a human. The large image on the left is the resimulated
    view. The dashed dark blue line indicates the ground truth lane
    center. The dashed light blue line represents the path driven by
    the human driver. The white dashes is the predicted path from the
    neural network. The yellow dashes are the tire positions from the
    resimulation. The red and green lines are the predicted lane
    edges.}
  \label{fig-aug-resim}
\end{figure}

In response to the challenges of creating realistic simulations, we
created the Augmented Resimulator tool, a solution that allows for
closed loop testing like in a synthetic simulator but working off real
sensor recordings instead of synthetic data. As mentioned in the
preprocessing section, PilotNet utilizes viewpoint transforms to
expand the training data to domains not recorded through human driving
(Section~\ref{sec-viewpoint-transform}). We leverage the same strategy
to generate testing environments from collected
videos. Figure~\ref{fig-aug-resim} shows a screenshot of the Augmented
Resimulator.

The basic approach is similar to video-replay, except the system under
test is free to control the car as if it was in a synthetic
simulation. At each new state of simulation, we produce sensor data
for the cameras through a viewpoint transform from the closest frame
in the recording. As long as the system under test doesn't deviate too
much from the recorded path, we will always have sensor data
available. If the network deviates too much from the recorded path,
then we will not have sufficient sensor information available to apply
our transformations; therefore in these instances we reset the
simulated vehicle to the center of the road. Of course, if we deviate
too far, we also consider this to be a failure. With this tool, we use
data collected from real-world cameras so that we do not have to
re-create the world photo-realistically in simulation.

Our approach is not without limitations, namely, the presence of image
artifacts as mentioned in Section~\ref{sec-viewpoint-transform}.
Furthermore, our driving scenarios are limited to data we recorded. We
cannot, for example, change the time of day on the fly as is possible
in synthetic simulation, nor can we take any turns or highway exits
unless they were recorded. We do, however, gain the advantage of
having as much simulated data as we can collect and label without the
need to design simulated cities and roads. Furthermore we can
accurately reproduce the exact scenario of real-world failures.

\subsection{Detecting Failures}
One metric that we are particularly interested in is the \gls{mdbf}
(see Section~\ref{sec-perf-history} for more information). We define
\gls{mdbf} as the distance driven under test, divided by the number of
failures in that distance. A naive criterion for failure is to detect
if there is a large lateral deviation from the human-driven
trajectory. However, this approach can fail when the road becomes too
wide, or if the human-driven path is not in the center of the road. To
better facilitate failure detection, we add additional labels per
frame in our test set recordings. These labels indicate the locations
of the left and right lane boundaries. If any of the wheels of the
simulated car touches these lane boundaries, we flag this as a
failure.

\subsection{In-Car Monitor}
\begin{figure}[htb]
  \hfil
  \includegraphics[width=\textwidth]{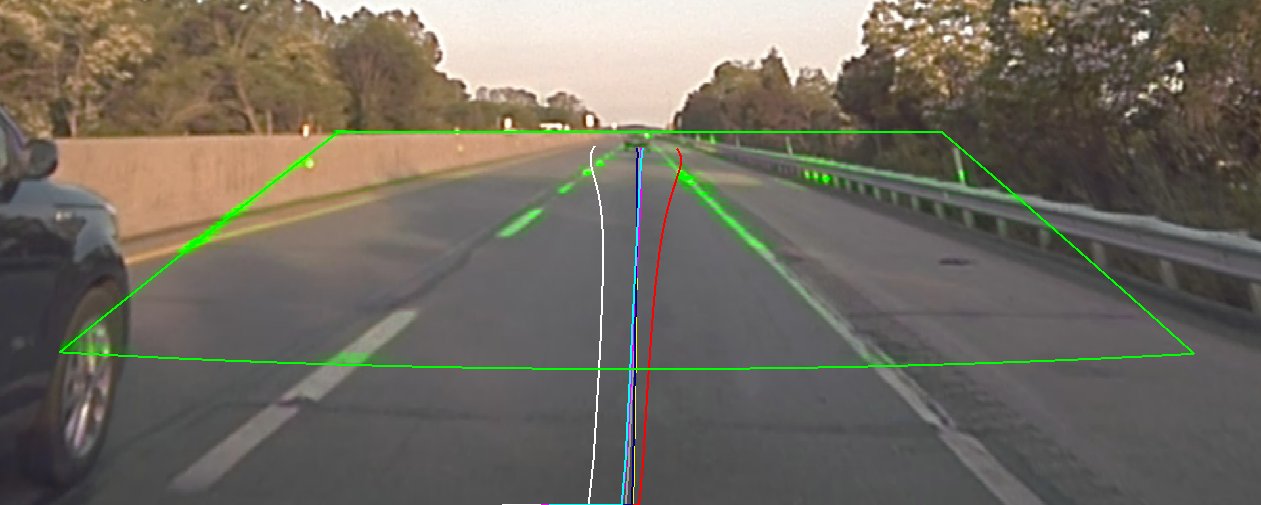}
  \caption{PilotNet in-car monitor showing the \gls{roi} (green
    trapezoid) and the seven predicted trajectories. The reason the
    trapezoid edges appear curved is that inference occurs on
    rectified images while the car display shows the original
    unrectified camera image. The straight line of the trapezoid in
    the rectified image therefore become slightly curved in the
    display image.}
  \label{fig-incar-monitor}
\end{figure}

We have created an in-car monitor to aid in system development. The
monitor, showing PilotNet's inputs and outputs, gives humans a
real-time view of PilotNet's performance. See
Figure~\ref{fig-incar-monitor}. A key feature of the monitor is the
several trajectories predicted by PilotNet for different maneuvers. In
the image, these trajectories are 100\m\ long.

\subsection{Where Does the Network Look?}
\label{sec-look-where}
\begin{figure}[p]
  \hfil
  \includegraphics[width=\textwidth]{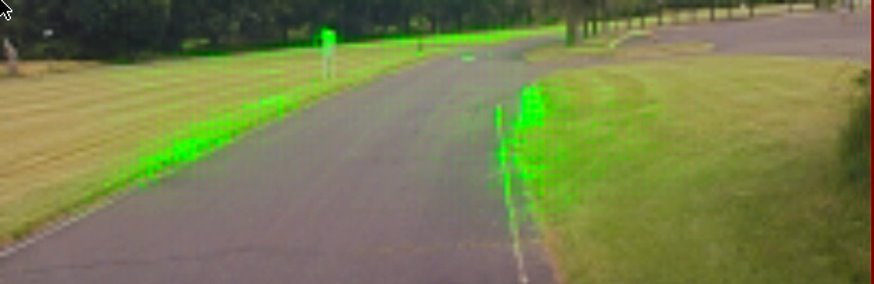}

  \hfil
  \includegraphics[width=\textwidth]{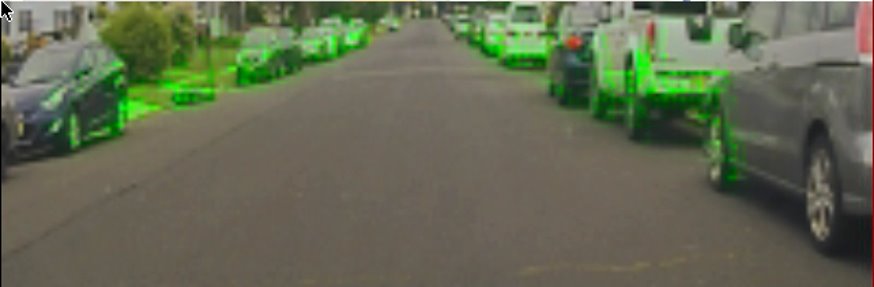}

  \hfil
  \includegraphics[width=\textwidth]{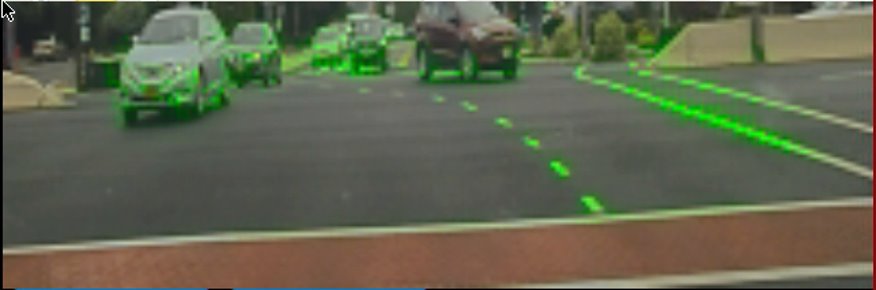}

  \hfil
  \includegraphics[width=\textwidth]{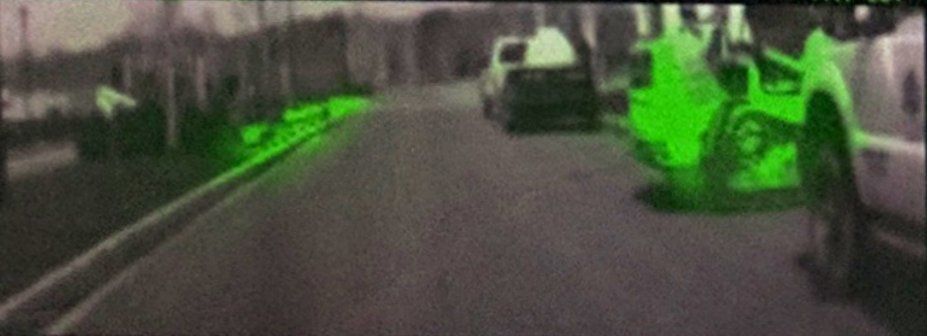}
  \caption{PilotNet saliency maps. The top image shows a narrow road
    with no lane markers. The network focuses on the grass-pavement
    boundary. The next lower image shows a residential street with no
    lane markings. Here the lower parts of the parked cars guide
    PilotNet. The third image from the top shows a busy intersection.
    Pilot pays attention to painted lines in the direction of travel,
    but ignores lines of the crosswalk that are nearly perpendicular
    to the direction of travel. In the bottom image PilotNet is guided
    by the BobCat front loader on the right and the curb on the left.}
  \label{fig-incar-saliency-vis}
\end{figure}

The in-car monitor includes a saliency map that highlights (in bright
green) the regions of the input image that are most salient in
determining PilotNet's output. The methodology in creating this
visualization is described in Bojarski et
al~\cite{explain-e2e-2017}. Figure~\ref{fig-pn-outputs} demonstrates
how the saliency map is computed. Figure~\ref{fig-incar-saliency-vis}
shows some examples of saliency maps.

\section{Metrics and Performance History}
\label{sec-perf-history}
\begin{figure}[htb]
  \hfil
  \includegraphics[width=\textwidth]{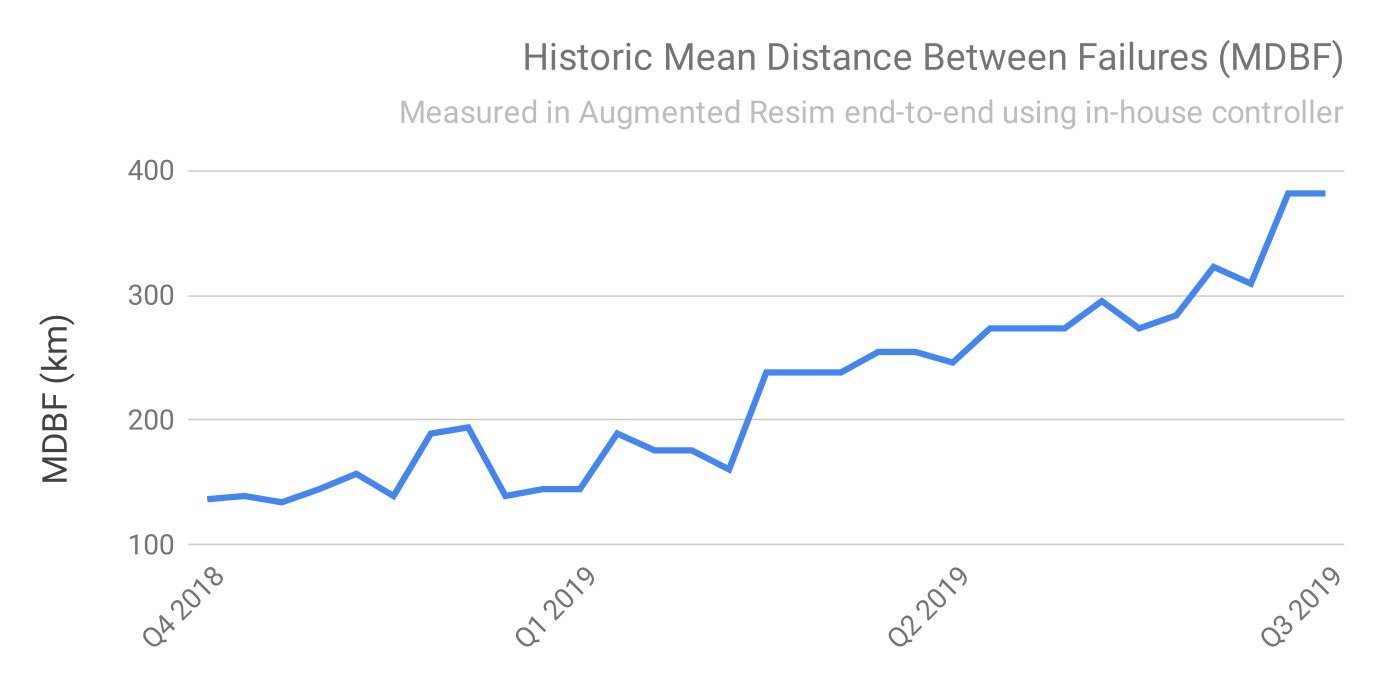}
  \caption{Evolution of PilotNet's \gls{mdbf} over time. Note that the
    \gls{mdbf} reflects the combination of the neural network and
    controller performance. Both are improved independently which can
    lead to temporary regressions of the overall driving
    performance. For example, as the controller was improved to more
    faithfully follow the predicted trajectory, shortcomings of
    PilotNet's predictions were uncovered.}
  \label{fig-incar-perf-history}
\end{figure}

A number of metrics are used to evaluate PilotNet performance. Most
critical is \gls{mdbf}. \gls{mdbf} measures the average distance
PilotNet can steer an autonomous vehicle before a human must take
control to avoid a dangerous situation. \gls{mdbf} is evaluated in
specific domains. Currently we are focusing on highway driving, which
means stretches of divided, limited-access roads, such as US
interstate highways. We exclude places where the highways split or
merge and where navigation instructions are more complex than just
``follow the road.'' In on-the-road tests, PilotNet was able to steer
from Holmdel, New Jersey to North Carolina and back with only one
highway intervention over a distance of about
1,300\km. (\href{https://developer.download.nvidia.com/video/PilotNet_1080p.mp4}{video
  NC to NJ}). We note that the conditions on this trip were not ideal,
with stretches of heavy rain and nighttime
driving. Figure~\ref{fig-incar-perf-history} shows the evolution of
\gls{mdbf}.

Another metric is ``precision'' or how closely the vehicle tracks to
the lane center. Our definition of precision is 100\x(1\m\ - RMS
deviation from lane center in meters) / 1\m. Right now we can only
measure precision for PilotNet in simulation (see
section~\ref{sec-aug-resim}). Typical values are 80\%. We find that
for human drivers precision is typically 66\%.

The final metric we track is ``comfort,'' which is a measure of the
``smoothness'' of the ride as measured by sensors on the vehicle. As
we measure it, comfort is determined by the \gls{rms} of the time
derivative of the vehicle's lateral acceleration. The comfort scale is
somewhat arbitrary, but we set constant acceleration to correspond to
a comfort of 100 and have comfort decrease in proportion to the
\gls{rms} time derivative of the vehicle acceleration. Human-driven
vehicles typically have comfort scores around 80. We achieve similar
scores in simulation with the current version of PilotNet. We hope to
measure PilotNet on-the-road comfort soon. One limitation of this
metric is that the comfort score drops when the vehicle enters a curve
(due to the increased lateral acceleration), even though the vehicle
may be driving perfectly in the center of the lane.

\section{Multi-Resolution Image Patches}
\label{sec-multires}

\subsection{Introduction}
Learning an accurate trajectory requires high-resolution data that
provides enough information for PilotNet to see clearly at a
distance. High-resolution cameras are used for this purpose. However,
the image preprocessing steps, such as cropping, and downsampling
significantly reduce resolution. In the early versions, the
preprocessed image patch fed into PilotNet had a size of 209x65
pixels. One of the advantages of a small patch size is that it reduces
the computing load, allowing PilotNet to drive with a high frame
rate. But the resolution (about 3.9 pixels per horizontal degree and
about 5.9 pixels per vertical degree) at large distances is too low
for the network to extract needed road features. To illustrate, at
this resolution, the full moon would be only 2 pixels wide.

\begin{figure}[htb]
  \hfil
  \begin{subfigure}{0.4\textwidth}
    \hfil
    \includegraphics[width=\linewidth]{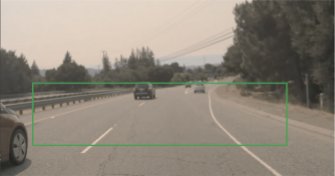}
    \caption{Original image}
    \label{fig-rect-roi-a}
  \end{subfigure}
  \hspace{1em}
  \begin{subfigure}{0.4\textwidth}
    \hfil
    \includegraphics[width=\linewidth]{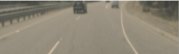}
    \caption{Regular patch}
    \label{fig-rect-roi-b}
  \end{subfigure}
  \caption{The original pinhole image is shown in
    \ref{fig-rect-roi-a} with the green rectangle indicating the
    \gls{roi}. The image in ~\ref{fig-rect-roi-b} shows the
    downsampled 209\x65 image. Note how few pixels are dedicated to
    the off ramp.}
  \label{fig-rect-roi}
\end{figure}

Figure~\ref{fig-rect-roi-a} shows an example of an original image from
the camera and Figure~\ref{fig-rect-roi-b} shows the processed image
patch. Note that in the patch, the off-ramp in the road ahead is
barely visible. This low resolution can confuse PilotNet, with the
predicted path swerving right before the off-ramp. One of the
solutions to this problem is to increase the patch resolution at large
distances. The most straightforward method is to uniformly increase
the patch resolution by reducing downsampling. However, uniformly
increasing the patch resolution will quadratically increase the
computational burden. For example, increasing the resolution five
times from 209\x65 to 1045\x325 requires 25 times more computation in
neural network training and inference.

To increase the resolution at far distances while keeping a near
constant computational load, we introduce a new method we call
``multi-resolution image patch.'' This method significantly improves the
PilotNet performance with modestly more computation, allowing PilotNet
to increase its \gls{mdbf} by about 50\%

\subsection{Creating the Image Patch}
\begin{figure}[htb]
  \hfil
  \includegraphics[width=\textwidth]{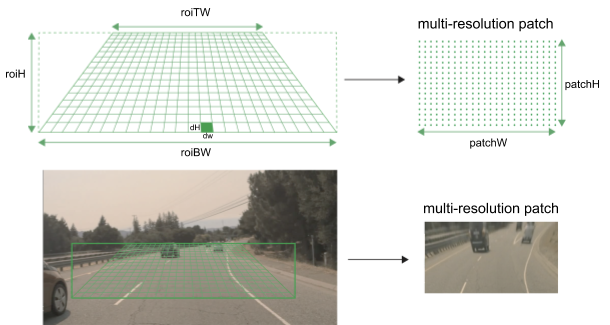}
  \caption{The \gls{roi} for the multi-resolution patch is a
    trapezoid. It is downsampled to generate a multi-resolution patch
    as shown on the bottom right.}
  \label{fig-multires-roi}
\end{figure}

The basic idea of the multi-resolution patch is to linearly increase
the horizontal and vertical resolutions (pixels per degree) from near
to far distance with respect to the regular image patch. The method we
used here extracts a trapezoidal ROI from the original image and then
reshapes it into a rectangular patch. Figure~\ref{fig-multires-roi}
shows how the multi-resolution patch is generated. Each pixel in the
image patch corresponds to a source area in the original image, and
the pixel value is the average value of the source area.

Let's denote the top width, the bottom width and the height of the
\gls{roi} in the original image as $roiTW$, $roiBW$ and $roiH$, and
the width and height of the image patch as $patchW$ and $patchH$. The
bottom width and height of the source area are denoted as $dW$ and
$dH$. For the multi-resolution patch, in pixel space, the width and
height of the source area are linearly increasing from top to
bottom. Here, we use a horizontal resolution ratio ($ratioW$) and a
vertical resolution ratio ($ratioH$) to define the source areas. The
horizontal/vertical resolution ratio is defined as the width/height
ratio of the bottom source area to the top source area. The
mathematical definition is as follows.

\begin{align}
  dW(i)&=a_w{}*i+b_w,    & i=0, 1, \cdots, patchW&-1 \\
  dH(i)&=a_h{}*i+b_h,    & i=0, 1, \cdots, patchH&-1
\end{align}
With constraints:
\begin{align}
  dW(patchH-1)            &=roiBW/patchW       \\
  ratioW                  &=dW(patchH-1)/dW(0) \\
  ratioH                  &=dH(patchH-1)/dH(0) \\
  \sum_{i=0}^{patchH-1}dH(i) &= roiH
\end{align}
where $i$ represents a row index starting at $0$ for the topmost row
of source areas. Hence, given $roiBW$, $roiH$, $patchW$, $patchH$,
$ratioW$ and $ratioH$, we can calculate the size of all source areas
using the above equations.

After defining the multi-resolution source areas, we downsample each
source area to one pixel by averaging the original pixel values,
obtaining the multi-resolution patch. Note that setting $ratioW=1$ and
$ratioH=1$, the generated patch is the same as the original patch.

\subsection{Experimental Results}
To evaluate the performance of the multi-resolution patch, we
performed a parameter search on $ratioW$ and $ratioH$ with fixed
$roiBW$, $roiH$, $patchW$, $patchH$ and $dH(patchH-1)$. Holding
$dH(patchH-1)$ constant ensures that the resolution at the bottom of
the resampled image remains fixed.

The dataset for the parameter search contained 20 hours of training
data and 14 hours of testing data. The optimal parameters reported by
the parameter search were further validated using a larger dataset
(260 hours of training data and 80 hours of testing data). The optimal
settings for PilotNet is $ratioW=2$ and $ratioH=8$ with a patch size
of 209\x113. That is, the horizontal and vertical resolution of the
top pixels of the multi-resolution patch are 2\x\ and 8\x\ larger than
those of the regular patch, respectively, which only increases the
computational load by 70\%.

\begin{figure}[htb]
  \hfil
  \includegraphics[width=\textwidth]{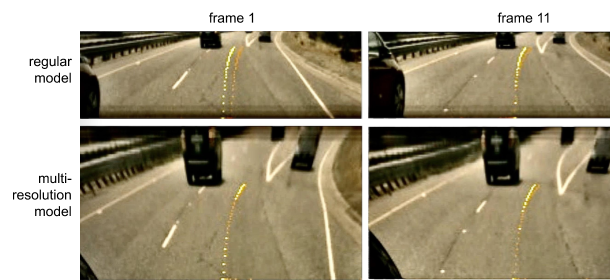}
  \caption{The yellow lines are ground truth trajectories, and the
    orange lines are the network-predicted trajectories. The
    multi-resolution PilotNet neural network achieves higher
    prediction accuracy. These are two example frames near forks.}
  \label{fig-multires-patches}
\end{figure}

The multi-resolution network (\ie, the network trained with the
multi-resolution patch) achieves a significantly higher performance in
the Augmented Resimulator (Section~\ref{sec-aug-resim}) both in terms
of \gls{mdbf} (increased by 50\%), comfort (increased by 3\%), and
precision (increased by 2\%). Multi-resolution also helps PilotNet
avoid swerving at off-ramps and other scenarios where lanes
split. Figure~\ref{fig-multires-patches} shows some example patches
where the multi-resolution network predicts a better driving
trajectory than the regular network.

We also tested the multi-resolution network in the car and compared it
with the regular network. Results were consistent with resimulation
results; the multi-resolution network drove smoothly at splits without
swerving (unlike the fixed-resolution network).

Multi-resolution image patches are now used in PilotNet by
default. This method increases the resolution of far-distance pixels
with slightly more computation in network training and inference. The
method is much less computationally expensive than simply increasing
the patch resolution with the same aspect ratio.

Note that the multi-resolution patch is “distorted” with respect to
the regular patch. However, this distortion is not a problem for
PilotNet since the network is able to learn the relationship between
the multi-resolution image space and 3D world space.

\section{Tuning the System}
\label{sec-tuning}

\subsection{A Detective Story: Catching a Network that Cheats}
As new versions of PilotNet were produced, performance in the Augmented
Resimulator tracked real on-the-road performance so that Resimulator
performance was assumed to be a reliable indicator of what we would see
in a real car.

In mid 2019 with increased training data, with tweaks to the network
architecture, and with adjustments in data sampling, \gls{mdbf} as
measured in the Resimulator continued to increase, but to our shock,
performance on the road became less stable. What was going on?

We got a hint that something was amiss when we saw in resimulation that
networks trained to only stay in lane did perfect lane changes. In these
instances, video sequences that captured lane changes were accidentally
included in the resimulation tests. After much thought we realized that
the network was inadvertently trained to generate paths that minimize
perturbation artifacts of the training images. Somehow the network was
keying on artifacts generated by the perturbations. In effect, the network
was getting high marks in resimulation by ”cheating”: it was using
resimulation artifacts to define the vehicle path rather than actual features
of the road. We realized that our efforts were essentially human-engineered
gradient descent in network hyperparameters space seeking to maximize
resimulation \gls{mdbf}.

Where do the artifacts come from? Recall that the Augmented
Resimulator produces simulated sensor images that are derived from
real-world recordings.  Since the simulated vehicle may drive a
different path than the vehicle in the real-world recording, the
real-world and simulated sensor positions will differ.  Therefore, a
transformation must be applied to the real-world sensor data in order
to ``simulate'' the alternate state that would be observed by the
sensors in the Augmented Resimulator. For example, if the simulated
vehicle tracks towards the right with respect to the real-world car,
the camera sensor generated image should show the view from the
simulated car as if it is driving closer to the edge of the right
lane. As we discussed in Section~\ref{sec-training-data-aug}, our
transformations make a flat-world assumption; any real objects that
stick up above the ground plane will have distortion artifacts. As
examples, look again at Figures~\ref{fig-shift-rot} and
\ref{fig-artifacts}.

We found it is possible that a network trained with the same
perturbation transformations as used in the Resimulator will utilize
the perturbations to gain an artificial improvement in Resimulator
metrics (\ie\ the network learns to cheat). This effect has been named
\gls{mapa}. A network that is affected by \gls{mapa} has a tendency to
follow the real-world human driving in resimulation, scoring good
metrics. Even though the network may have good scores in Augmented
Resimulation, it will drive poorly in the real world, where these
augmentation artifacts are not present. We note that cheating is even
observed on synthetic flat-world images. It appears that networks can
learn to cheat not only on clearly distorted images like lampposts,
but even on subtle pixel-level effects not visible to humans.

\subsection{Measuring if the Network is Affected by Augmentation Artifacts}
In order to examine whether a network is using augmentation artifacts
to gain an artificially inflated score in the Augmented Resimulator, a
specific test was created. Data was collected on the New Jersey Garden
State Parkway (NJ-GSP) by driving the route twice, first with the
human driving close to the left edge of the lane, and later, with the
human driving close to the right edge of the same lane.  Two separate
Resimulations were executed, one using the left-biased recording, and
the other using the right-biased recording.

If the network is not affected by \gls{mapa}, it is expected that the
network will drive in the center of the lane in both Resimulations. If
the network is affected by \gls{mapa}, it is expected that it will be biased
in the same direction as the human.

\subsubsection{\gls{mapa} Score}
We created a formula using the mean values of bias in this test to
compute a \gls{mapa} score. We wanted this score to have the following
properties:
\begin{itemize}
  \item The score should give 0\% for networks not affected by
  \gls{mapa} (zero affinity).
  \item The score should give 100\% if the network drives exactly
  like the human (note that it is possible to go above 100\% if the bias is even
  greater than that of the human).
  \item The score should be unaffected by network bias, so a network
    that tracks too far to the left but is unaffected by augmentation
    artifacts should still report 0\%. Having 0\% \gls{mapa}
    coefficient does not mean the network drives in the center of the
    road or even drives well.
\end{itemize}

We created a formula for this score that satisfies the above conditions:
\begin{equation}
  \textrm{\gls{mapa} score} = \frac{1}{2}\left|\frac{y_{L} - y_{average}}{y_{HL}} +
    \frac{y_{R} - y_{average}}{y_{HR}}\right| \times 100\%
\end{equation}
where, $y_L$ is the average lateral offset in meters from the center
of the lane for the resimulation using the left-biased recording,

$y_R$ is the average lateral offset in meters from the center of the
lane for the resimulation using the right-biased recording,

$y_{average}$ is the average of $y_L$ and $y_R$, and

$y_{HL}$ and $y_{HR}$ are the average offset by the human driver in
the left and right recording respectively.

For all these parameters, a left deviation from center has a positive
value and a right deviation has a negative value.

We use this formula and test to detect when networks are
cheating. Networks that exhibit poor \gls{mapa} scores will have
resimulation results that are inconsistent with testing on a real car.

As an example, suppose $y_L = 0.5$, $y_R = -0.5$, $y_{HL} = 1$, and
$y_{HR} = -1$.  Then, $y_{average} = 0$.

In this case,
\begin{equation*}
  \textrm{\gls{mapa} score} = \frac{1}{2}\left|\frac{0.5 - 0}{1} + \frac{-0.5 - 0}{-1}\right|
  \times 100\% = \frac{1}{2}(0.5 + 0.5) \times 100\%  = 50\%.
\end{equation*}
With networks trained with human trajectories as ground truth and
optimized to get a high \gls{mdbf} in resimulation we often suffered
\gls{mapa} scores above 50\%. To reduce \gls{mapa} scores we settled
on a different way to train our networks: we used human-created labels
of the lane centers as ground truth. The inherent random departures
from the lane centers by the human data collection drivers resulted in
a randomization of the artifacts. Networks trained with lane-center as
ground truth had \gls{mapa} scores less than 5\%. On the road testing
of these lane-center trained networks showed excellent performance.

\section{Lessons Learned}
\label{sec-lessons}

\subsection{Diagnostic Tools Are More Important than Network Architecture}
Data quality is key to training large \glspl{dnn}. No amount of
network architecture tuning can overcome bad training data (\eg\ data
with incorrect labels).

We observed that the path predicted by PilotNet would sometimes jiggle
back and forth from one frame to the next. While our test car still
drove reasonably well, acting as a low pass filter, overall
performance suffered. We eventually traced the jitter to mislabeled
training data. When the bad labels were removed from the training
data, the predicted path became more stable and the performance
metrics improved. One of our data cleaning tools is called the ``bad
apples tool.'' That name derives from the American saying that ``one
bad apple spoils the whole barrel (of apples)'', and reflects our
observation that even a small number of corrupt training examples have
a disproportionate effect on performance.

Given the crucial nature of establishing a huge, clean training data
corpus, efficient and accurate tools are essential. Initially we did
not perform sufficient tests at each of the stages in data collection.
We now recognize the necessity of maintaining data integrity at every
stage.

\subsection{Use Independent Ways to Validate Results}
As an example, we created software to do viewpoint transforms to
correct for different camera placements. To be sure the transforms
were working correctly we built a tool where we could move a camera
from one known position to another and then apply the transform for
images taken at each position. We then subtracted one image from the
other creating a difference image. We knew the transform was working
when the difference image had pixel values near zero.

\subsection{Debugging and Visualization Tools Are Important
  Because \glspl{dnn} Tend to Hide Bugs}

One of the virtues of neural networks is that they are relatively
insensitive to small changes in network weights. A fair number of
weights can be set to zero and the network will usually still provide
a usable output. The flip side of this robustness is that errors in
architecting or training the network will often yield apparently
functioning systems. However, these systems will not provide peak
performance. In mission critical applications like driving where near
perfect performance is required, any reduction in performance can have
fatal consequences. Therefore it is critical to have some tools to
detect hidden bugs. One tool we use is the visualization tools
described in Section~\ref{sec-look-where}. Another tool is the
examination of ``learning curves.'' These are plots of training loss
and test loss as a function of training set size. We can also do
similar plots of \gls{mdbf} on a test set as a function of training
set size. In either case, loss and \gls{mdbf} should improve as the
training set becomes larger. In practice, we found this is not always
the case. We have seen that performance often improves up to some
limit, and then flattens. While this could be due to insufficient
network capacity, this possibility is easily checked by increasing
network size. Most strikingly, we observed that as we uncovered bugs,
performance again improves until it flattens at some higher level. We
have found that this performance saturation is often caused by corrupt
training data, prompting us to focus on data integrity and data
cleaning.

\subsection{Automated Unit Tests Are a Must}
Given the complexity of developing Autonomous Vehicles, teams will
inevitably grow and become specialized. As this happens, it's
important to update development infrastructure and processes to keep
the team productive.

A small team (1--10 developers) can get by without a \gls{ci}
infrastructure or code review barriers; developers can run integration
and unit tests themselves on an honor system before merging code, and
code review can be done verbally after merge. Shouting ``I'm merging
the big change now!'' is effective in a group this small if it's
co-located, and absent team members can be briefed when they return.

Manual \gls{ci} and post-merge review do not scale to a medium-sized
team (10–50 developers) or a team that is not co-located which will
inevitably happen when team members are absent. Without automated unit
testing pre-merge, the code base will regress, and in a medium-sized
team, the cost and frequency of regressions is high enough to stifle
progress. Code reviews in a medium-sized team serve the important
purpose of distributing knowledge about the code. Reviewers become
aware of changes before they happen and can apply their knowledge to
keep the code in a high-functioning state rather than fixing technical
debt after the fact. Code reviews replace verbal communication as the
communication tool to talk about code and a reference for new team
members.

When scaling to a large team (50–-200 developers), automated \gls{ci}
and code review become critical. Teams without automated CI will be
unable to make progress and will find themselves fighting the fires of
regression more than they spend on actual work. At this size, a
dedicated \gls{ci} team will likely be required to keep developers
productive. Teams without code review will find their previously slim
architecture growing into an unmanageable pile of code patched to
correct the latest bug without consideration for long-term vision. The
volume of code reviews makes them unsuitable for distributing
information, and formal feature and requirement planning replace code
review as a method to broadcast the important changes happening to the
code.

\section{Conclusions}
\label{sec-conclusions}
In this document we described an experimental research system,
PilotNet, and presented a key performance metric and its improvement
of the years, We also described some of the numerous aspects of the
data collection, neural network training, and simulation analysis that
let PilotNet evolve from promising demo to major component of a system
that can drive long distances in challenging conditions without human
intervention. The purpose of PilotNet was to gain valuable insight
into the nature of the immense AV challenges and potential
solutions. Actual production systems are far more complex than
PilotNet and include diversity and redundancy for safety.

\section*{Videos}
\begin{compactenum}
  \setlength{\itemsep}{3pt}
  \item\href{https://cs.nyu.edu/~yann/research/dave/}{\gls{dave} robot
    off-road obstacle avoidance} (videos and images)
  \item\href{https://youtu.be/qhUvQiKec2U}{PilotNet demo at \gls{gtc}
    San Jose 2016} (Youtube)
  \item\href{https://www.youtube.com/watch?v=-96BEoXJMs0}{PilotNet
    demo at \gls{gtc} Europe 2016} (Youtube)
  \item\href{https://www.youtube.com/watch?v=FABftuXUOxE}{\gls{ces}
    live demo coverage by The Verge} (Youtube)
  \item\href{https://www.youtube.com/watch?v=mCmO_5ZxdvE}{PilotNet on
    Lombard Street} (Youtube)
  \item\href{https://developer.download.nvidia.com/video/PilotNet_1080p.mp4}{Highway
    test drive from North Carolina to New Jersey} (NVIDIA Developer
    Zone)
\end{compactenum}

%
%
\small
\bibliography{neural-nets,pilotnet}

\begin{thebibliography}{10}

\bibitem{lecun-89e}
Y.~LeCun, B.~Boser, J.~S. Denker, D.~Henderson, R.~E. Howard, W.~Hubbard, and
  L.~D. Jackel.
\newblock Backpropagation applied to handwritten zip code recognition.
\newblock {\em Neural Computation}, 1(4):541--551, Winter 1989.
\newblock URL: \url{http://yann.lecun.org/exdb/publis/pdf/lecun-89e.pdf}.

\bibitem{LMB*05}
Yann LeCun, Urs Muller, Jan Ben, Eric Cosatto, and Beat Flepp.
\newblock Off-road obstacle avoidance through end-to-end learning.
\newblock In {\em Proceedings of Advances in Neural Information Processing
  Systems NIPS*04}. MIT Press, 2005.
\newblock URL:
  \url{https://papers.nips.cc/paper/2847-off-road-obstacle-avoidance-through-end-to-end-learning}.

\bibitem{Pom89}
Dean Pomerleau.
\newblock {ALVINN}: An autonomous land vehicle in a neural network.
\newblock In D.S. Touretzky, editor, {\em Proceedings of Advances in Neural
  Information Processing Systems 1 (NIPS*88)}, pages 305--313. Morgan Kaufmann,
  January 1989.
\newblock URL:
  \url{https://papers.nips.cc/paper/95-alvinn-an-autonomous-land-vehicle-in-a-neural-network.pdf}.

\bibitem{JHK*07}
L.D. Jackel, Douglas Hackett, Eric Krotkov, Michael Perschbacher, James
  Pippine, and Charles Sullivan.
\newblock How darpa structures its robotics programs to improve locomotion and
  navigation.
\newblock {\em Communications of the {ACM}}, 50, November 2007.
\newblock URL:
  \url{https://cs.uwaterloo.ca/~brecht/courses/854-Experimental-Performance-Evaluation-2018/readings/darpa-robotics-cacm-2007.pdf}.

\bibitem{dave2-2016}
Mariusz Bojarski, Davide~Del Testa, Daniel Dworakowski, Bernhard Firner, Beat
  Flepp, Prasoon Goyal, Lawrence~D. Jackel, Mathew Monfort, Urs Muller, Jiakai
  Zhang, Xin Zhang, Jake Zhao, and Karol Zieba.
\newblock End to end learning for self-driving cars, April 25 2016.
\newblock URL: \url{http://arxiv.org/abs/1604.07316}, \href
  {http://arxiv.org/abs/arXiv:1604.07316} {\path{arXiv:arXiv:1604.07316}}.

\bibitem{husky}
Clearpath Robotics.
\newblock Husky unmanned ground vehicle.
\newblock URL:
  \url{https://clearpathrobotics.com/husky-unmanned-ground-vehicle-robot/}.

\bibitem{hyperion7}
{NVIDIA}.
\newblock {NVIDIA DRIVE Hyperion\texttrademark\ Developer Kit}.
\newblock URL: \url{https://developer.nvidia.com/drive/drive-hyperion}.

\bibitem{nvdrive}
{NVIDIA}.
\newblock {NVIDIA DRIVE\texttrademark\ -- Software}.
\newblock URL: \url{https://developer.nvidia.com/drive/drive-software}.

\bibitem{pmlr-v9-ross10a}
Stephane Ross and Drew Bagnell.
\newblock Efficient reductions for imitation learning.
\newblock In Yee~Whye Teh and Mike Titterington, editors, {\em Proceedings of
  the Thirteenth International Conference on Artificial Intelligence and
  Statistics}, volume~9 of {\em Proceedings of Machine Learning Research},
  pages 661--668, Chia Laguna Resort, Sardinia, Italy, 13--15 May 2010. PMLR.
\newblock URL: \url{http://proceedings.mlr.press/v9/ross10a.html}.

\bibitem{NIPS2019_9343}
Pim de~Haan, Dinesh Jayaraman, and Sergey Levine.
\newblock Causal confusion in imitation learning.
\newblock In H.~Wallach, H.~Larochelle, A.~Beygelzimer, F.~d'~Alch\'{e}-Buc,
  E.~Fox, and R.~Garnett, editors, {\em Advances in Neural Information
  Processing Systems 32}, pages 11698--11709. Curran Associates, Inc., 2019.
\newblock URL:
  \url{http://papers.nips.cc/paper/9343-causal-confusion-in-imitation-learning.pdf}.

\bibitem{1512.03385}
Kaiming He, Xiangyu Zhang, Shaoqing Ren, and Jian Sun.
\newblock Deep residual learning for image recognition, 2015.
\newblock URL: \url{https://arxiv.org/abs/1512.03385}, \href
  {http://arxiv.org/abs/arXiv:1512.03385} {\path{arXiv:arXiv:1512.03385}}.

\bibitem{explain-e2e-2017}
Mariusz Bojarski, Philip Yeres, Anna Choromanska, Krzysztof Choromanski,
  Bernhard Firner, Lawrence Jackel, and Urs Muller.
\newblock Explaining how a deep neural network trained with end-to-end learning
  steers a car, 2017.
\newblock URL: \url{https://arxiv.org/abs/1704.07911}, \href
  {http://arxiv.org/abs/arXiv:1704.07911} {\path{arXiv:arXiv:1704.07911}}.

\end{thebibliography}
\bibliographystyle{unsrturl}
%
\label{lastpage}
\end{document}